\pdfoutput=1

\documentclass[11pt]{article}

\usepackage[preprint]{acl}
\usepackage{graphicx}
\usepackage{times}
\usepackage{latexsym}
\usepackage{amssymb}
\usepackage{booktabs}
\usepackage{amsmath}
\usepackage[T1]{fontenc}

\usepackage[utf8]{inputenc}

\usepackage{microtype}
\usepackage{graphicx}
\usepackage{inconsolata}
\usepackage{enumitem}
\usepackage{xspace}
\usepackage{microtype}
\newcommand{\specialsize}[1]{{\fontsize{12}{12}\selectfont #1}}
\newcommand\dataset{S{\small TICKER}C{\small ONV}\xspace}
\newcommand\ourmodel{PEGS\xspace}
\newcommand\prag{\ourmodel-RAG\xspace}
\newcommand\pgen{\ourmodel-Gen\xspace}
\newcommand\pret{\ourmodel-Ret\xspace}
\newcommand\fourmodel{\textbf{PE}rceive and \textbf{G}enerate \textbf{S}tickers\xspace}
\newcommand\agent{Agent4SC\xspace}
\newcommand\fagent{Agent for S{\small TICKER}C{\small ONV}\xspace}
\newcommand\github{https://github.com/ZhangYiqun018/StickerConv}
\newcommand\project{https://neu-datamining.github.io/StickerConv}

\setlist[itemize]{leftmargin=*}
\usepackage{svg}
\usepackage{mathtools}
\usepackage{hyperref}
\hypersetup{hypertex=true,
            colorlinks=true,
            linkcolor=red,
            anchorcolor=red,
            citecolor=blue}
\usepackage{bm}
\usepackage{subfigure}
\usepackage{subcaption}
\usepackage{stfloats}
\usepackage{color}
\usepackage{xcolor}
\usepackage{multirow}

\title{S\specialsize{TICKER}C\specialsize{ONV}: Generating Multimodal Empathetic Responses \\from Scratch}

\author{
 Yiqun~Zhang$^{1*}$, Fanheng~Kong$^{1*}$, Peidong~Wang$^{1*}$, Shuang~Sun$^{1}$, Lingshuai~Wang$^{1}$, \\  \textbf{Shi~Feng}$^{1\dag}$, \textbf{Daling~Wang}$^{1}$, \textbf{Yifei~Zhang}$^1$, \textbf{Kaisong~Song}$^2$ \hspace{0.4em} \\
 $^1$School of Computer Science and Engineering, Northeastern University, Shenyang, China \\
 $^2$Alibaba Group, Hangzhou, China  \\
\texttt{\{yiqunzhang, kongfanheng, pdongwang, shuangsun, lingshuaiwang\}@stumail.neu.edu.cn}\\
\texttt{\{fengshi, wangdaling, zhangyifei\}@cse.neu.edu.cn}\\
\texttt{kaisong.sks@alibaba-inc.com}
}

\begin{document}
\maketitle
\begin{abstract}
Stickers, while widely recognized for enhancing empathetic communication in online interactions, remain underexplored in current empathetic dialogue research, 
notably due to the challenge of a lack of comprehensive datasets.
In this paper, we introduce the \fagent (\agent), which uses collaborative agent interactions to realistically simulate human behavior with sticker usage, thereby enhancing multimodal empathetic communication. Building on this foundation, we develop a multimodal empathetic dialogue dataset, \dataset, comprising 12.9K dialogue sessions, 5.8K unique stickers, and 2K diverse conversational scenarios. 
This dataset serves as a benchmark for multimodal empathetic generation.
To advance further, we propose \fourmodel (\ourmodel), a multimodal empathetic response generation framework, complemented by a comprehensive set of empathy evaluation metrics based on LLM. Our experiments demonstrate \ourmodel's effectiveness in generating contextually relevant and emotionally resonant multimodal empathetic responses, contributing to the advancement of more nuanced and engaging empathetic dialogue systems\footnote{\url{\github}}.
\end{abstract}

\def\thefootnote{\dag}\footnotetext{Corresponding author.}\def\thefootnote{\arabic{footnote}}

\def\thefootnote{*}\footnotetext{Equal contribution.}\def\thefootnote{\arabic{footnote}}

\section{Introduction}

Increasing research indicates that utilizing stickers in online chats can effectively alleviate stress, augment personal happiness, and notably boost empathy \cite{karikoLaughingOneSelf2019,akramMentalHealthMemes2022,jiangReciprocityHomophilySocial2023}. Prior studies on stickers primarily concentrated on sentiment analysis \cite{ge-etal-2022-towards} and recommendation systems \cite{gaoLearningRespondStickers2020a, zhangSelectingStickersOpenDomain2022}, overlooking their vast potential in empathetic response generation. Most empathetic response generation tasks focus solely on textual modality \cite{fuECOREEmotionCorrelation2023, lee-etal-2022-gpt, jiangReciprocityHomophilySocial2023}, yet stickers in chats convey more abundant and intuitive emotional information, enhancing the expressiveness and emotional depth of responses.

\begin{figure}[t]
    \centering
    \includegraphics[width=\linewidth]{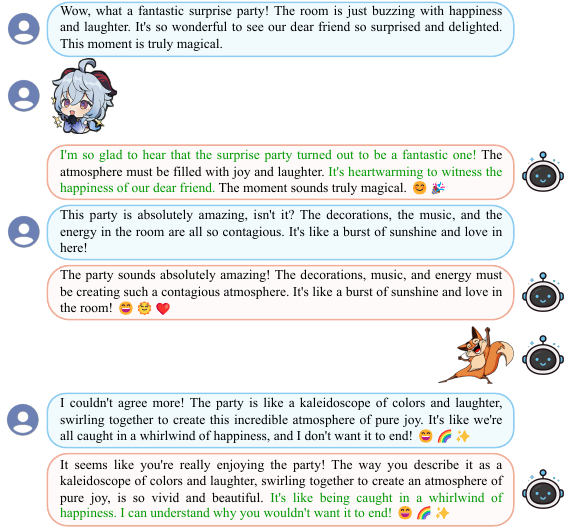}
    \caption{An example of multimodal conversation in the \dataset. Both parties can utilize the stickers to express their emotions, which enhances interactivity and expression. The assistant can empathize with the user according to the conversation (\textcolor[RGB]{0,153,0}{green} text).}
    \label{group_chat}
    \vspace{-1em}
\end{figure}

Integrating stickers with textual communication and interspersing stickers within the dialogue can yield more varied and superior-quality empathetic replies. A primary challenge in integrating stickers into empathetic response generation is developing a high-quality dataset to support this innovative multimodal communication. 
To address this, we leverage large language model (LLMs) for dataset construction.
LLMs, with their extensive world knowledge and text processing capabilities \citep{openai2023gpt, touvron2023llama, touvron2023llama2}, demonstrate near-human annotation abilities \cite{wangSelfInstructAligningLanguage2023, pangakisAutomatedAnnotationGenerative2023}. However, applying LLMs directly have limitations in empathetic tasks, excelling in responding to explicit human instructions but lacking proactivity, a critical aspect of empathy \cite{wuMultimodalLargeLanguage2023, yinSurveyMultimodalLarge2023}. Empathy necessitates understanding others' emotions and the ability to actively express support and understanding \cite{zechTalkingEmotionalExperience2005, sharmaComputationalApproachUnderstanding2020}. To mitigate this, we introduce a multi-agent system based on LLMs, \textbf{\fagent} (\agent). This system, through inter-agent interactions, utilizes stickers to simulate human-like dialogue scenarios. It not only generates text responses but also strategically selects suitable stickers, thereby effectively enhancing empathy.

Based on \agent, we build a multimodal empathetic dataset, \textbf{\dataset}, that comprises 12.9K dialogue sessions and 5.8K unique stickers. 
\dataset boasts an average of 5.22 stickers per dialogue session, mirroring the sticker usage patterns observed in human communication.
Figure \ref{group_chat} depicts an example of conversations in our dataset. To the best of our knowledge, this is the first multimodal empathetic dialogue dataset, with the particular utility of sticker as non-textual modal information to better facilitate empathy.

Although \agent effectively generates multimodal empathetic responses, it is limited by expensive inference costs and specific sticker databases. To further advance the research on multimodal empathetic dialogue, we develop an end-to-end multimodal empathetic response generation framework, \textbf{\ourmodel}, with the ability to \fourmodel. Beyond the general ability to generate textual empathetic responses, \ourmodel receives multimodal inputs and autonomously generates stickers based on the emotional and contextual aspects of the dialogue at the appropriate moment. Furthermore, to simulate human communications on social media in the real world, our model supports interleaved multiple image and text inputs.

Misalignments between the modal quantities of predicted and golden responses can distort evaluation outcomes, and empathy is difficult to quantify due to its subjective nature. To address this, we propose a novel method for evaluating multimodal empathetic responses, focusing on \textbf{empathy}, \textbf{consistency}, and \textbf{ranking}. Utilizing the extensive world knowledge and anthropomorphic abilities of LLMs, this approach provides solid support for assessing multimodal empathetic replies.

In conclusion, the main contributions of this work are as follows:

\begin{itemize}[nolistsep, noitemsep]
    \item We introduce an LLM-based multi-agent system, \fagent (\agent), which integrates stickers into empathetic dialogues, ensuring contextual consistency, variety, and empathy aligned with human interactions.  Using \agent, we create a multimodal empathetic dialogue dataset, \dataset.
    \item We design \fourmodel (\ourmodel), a multimodal empathetic response generation framework that intuitively incorporates stickers based on the emotional and contextual dynamics of the dialogue. \ourmodel adeptly processes multimodal inputs, generating empathetic textual responses and using stickers appropriately to enhance these responses.
    \item We propose a method for assessing multimodal empathetic responses. It leverages LLM to evaluate the quality of these responses, with a specific focus on empathy, consistency, and ranking.

\end{itemize}

\section{Related Work}
\subsection{Empathetic Response Generation}
Empathetic response generation focuses on enabling machines to understand and respond to human emotions. The foundational EMPATHETICDIALOGUES dataset \cite{rashkinEmpatheticOpendomainConversation2019a} and subsequent innovations like the empathetic listener mixture model \cite{linMoELMixtureEmpathetic2019a} have significantly advanced this area. Large Language Models (LLMs) like ChatGPT have been explored for empathetic response generation, although their application remains limited \cite{lee-etal-2022-gpt,zhou-etal-2023-facilitating}. Challenges persist, especially in leveraging multimodal information for richer emotional engagement and in accurately evaluating empathetic responses due to their subjective nature. These obstacles highlight the ongoing need for research in effectively integrating LLMs and multimodal data into empathetic dialogue. Additionally, the subjective nature of empathy complicates its quantitative assessment \cite{fuECOREEmotionCorrelation2023, lee-etal-2022-gpt}, posing a further obstacle to the field's advancement.

\subsection{Large Multimodal Models}
LLMs, such as ChatGPT \citep{openai2023gpt}, LLaMA \citep{touvron2023llama, touvron2023llama2}, demonstrate powerful capabilities in dialog interaction and instruction following, and recent researches have extended LLMs to multimodal domains. Flamingo \citep{alayrac2022flamingo} exhibits promising zero-shot and few-shot multimodal understanding capability by adding a cross-attention layer to connect the frozen vision encoder with the LLM. BLIP \citep{li2023blip, dai2023instructblip}, MiniGPT-4 \citep{zhu2023minigpt, chen2023minigptv2} and LLaVA \citep{liu2023visual, liuLlavav1ImprovedBaselines2023} bridge the frozen vision encoder and the LLM through a small intermediate model. \citet{kohGILLGeneratingImages2023} for the first time explore the mapping of the output of LLMs into the input space of the vision decoder, empowering LLMs with image generation capability. GILL \citep{kohGILLGeneratingImages2023}, MiniGPT-5 \citep{zhengMiniGPT5InterleavedVisionandLanguage2023} align LLMs to frozen vision decoders through an encoder-decoder transformer, while SEED \citep{ge2023planting} employs a learnable Q-Former. In contrast to previous works which favor realistic images, our target is to generate stickers, which are abstract and exhibit distinct emotional tendencies.

\subsection{LLM-Based Agents}
LLM-based agents mark a major leap in AI, leveraging their capabilities for tasks like reasoning and interaction, as shown in recent studies \cite{wangSurveyLargeLanguage2023a,liCAMELCommunicativeAgents2023}. They find uses across various domains, such as software engineering \cite{qianCommunicativeAgentsSoftware2023} and scientific inquiry \cite{boikoEmergentAutonomousScientific2023}, highlighting their versatility. These agents can imitate complex human actions, partake in social interactions \cite{parkGenerativeAgentsInteractive2023, tuCharacterChatLearningConversational2023}, and replicate intricate scenarios like elections \cite{argyleOutOneMany2022}, debates \cite{wangApolloOracleRetrievalAugmented2023}, and consumer patterns \cite{wangWhenLargeLanguage2023}, illustrating their capacity to emulate human social dynamics. A notable innovation is their use in generating AI training data. Studies \cite{wangSelfInstructAligningLanguage2023, pengGeneratingEfficientTraining2023, zhouCharacterGLMCustomizingChinese2023} highlight their efficiency and cost-effectiveness in producing high-quality training datasets, transforming AI model development.

\begin{figure*}[!htbp]
    \centering
    \includegraphics[width=0.9\linewidth]{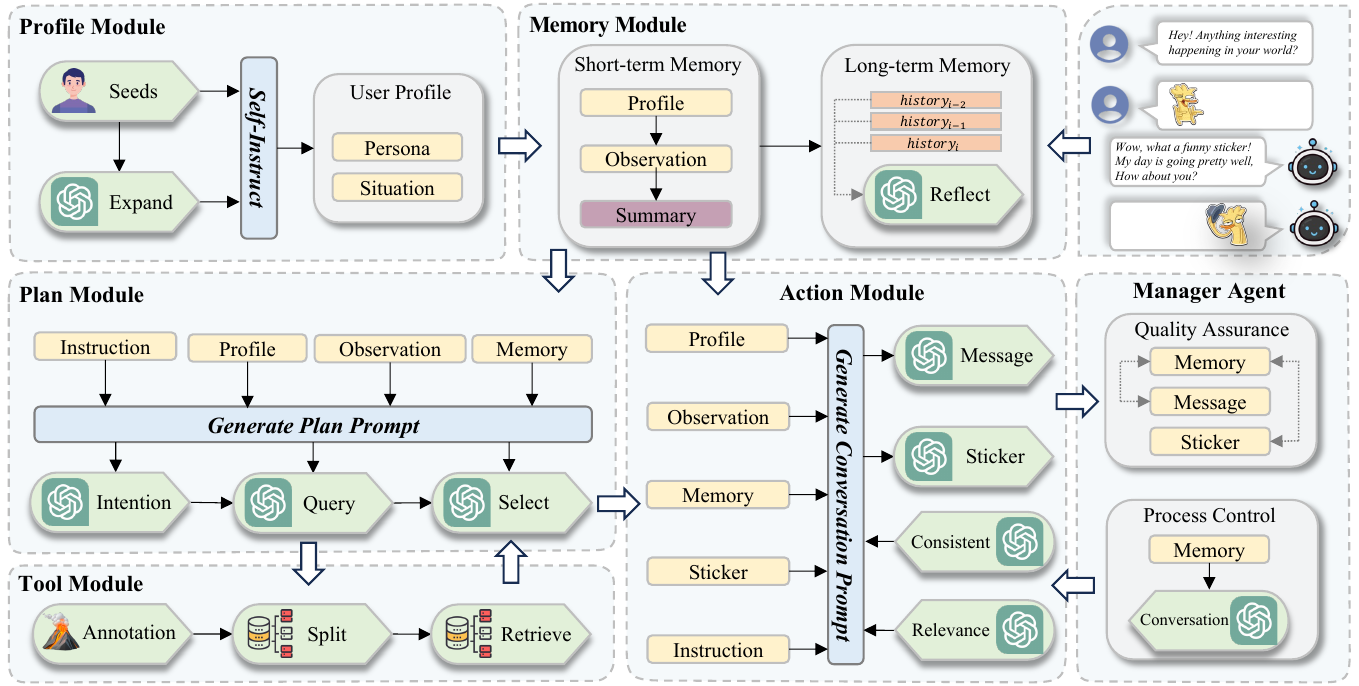}
    \caption{
The overview of \agent. Memory and Plan modules enable the agent to mimic human observation and thought, overcoming LLMs' inability to grasp nuanced emotions. The Action module supports generating insights with human-like emotional reactions. The Profile module gives each agent distinct reflections and actions. Furthermore, \agent uses stickers as a Tool for more natural conversation, allowing the agent to choose stickers like humans. These modules streamline observation, reflection, and action, while the Manager Agent maintains performance and quality.
}
    \vspace{-0.7em}
    \label{fig:agent_system}
\end{figure*}

\section{\fagent}
Confronted with the pivotal challenge of a lack of datasets for multimodal empathetic response tasks, we craft our own dataset utilizing large language models (LLMs). Nevertheless, LLMs experience difficulties in grasping nuanced human emotions and initiating actions beyond explicit directives. These limitations render LLMs and large multimodal models (LMMs) less proficient in the generation of multimodal empathetic responses. In response to these issues, we introduce \fagent (\agent), a multi-agent system predicated on LLM, devised to mimic human conversational patterns. 
Figure \ref{fig:agent_system} presents the overview of \agent. 
By integrating multiple modules and the strategic use of stickers, \agent aims to generate emotional and varied empathetic responses, thereby overcoming the inherent deficiencies of LLMs in empathetic engagements. 

\subsection{Profile Module}
\label{section:profile}
The profile module, comprising \textbf{Persona} and \textbf{Situation} components, underpins its foundation by defining users' personality traits and behavioral patterns for empathetic interactions. Persona outlines users' character traits, backgrounds, and experiences, while Situation details their current circumstances and emotional states. 

To enrich and diversify user profiles, we initially create profiles with varied emotional distributions, then expand to 2,000 unique profiles using \textbf{SELF-INSTRUCT} \cite{wangSelfInstructAligningLanguage2023} method. This approach results in a profile archive with a broad emotional spectrum, as shown in Figure \ref{fig:profile distribution}, aiming to enhance the system's ability to simulate human-like responses in empathetic dialogues.

\subsection{Tool Module}
To adapt the SER30K dataset, with its 1,887 themes and 30,739 emotion-tagged stickers \citep{liuSER30KLargeScaleDataset2022a}, for human-like sticker use in dialogues, we transform it into a tool. Integrating it into the agent system faced several hurdles: the inappropriateness of many stickers for empathetic dialogues, the lack of detailed content or emotional analysis in the single-emotion labeled stickers, and the contrast between the extensive SER30K collection and the limited, personalized collections humans use.

To make stickers an effective tool, we refine the SER30K through \textbf{filtering}, \textbf{annotating}, and \textbf{knowledge extraction} (detailed in Appendix \ref{app: tool_module}). 
Each sticker is enriched with three pieces of information: emotion, description, and recommendation. Subsequently, SER30K is segmented into smaller, emotionally balanced vector databases.
A critical action, "\textit{\textbf{Retrieve}}", is then designed for agents to efficiently use these databases by searching for stickers based on the communication context and desired emotion, with the system suggesting the top-K closest matches for selection.

\subsection{Memory Module}
In empathetic interactions, response quality hinges on communication coherence and personality trait consistency. Agents must learn and give feedback by observing during interactions and ensure long-term personality consistency. Following the Generative Agent \cite{parkGenerativeAgentsInteractive2023}, we use short-term updates of core traits from profiles and observations for immediate memory. For long-term memory, continuous interactions and self-reflection help maintain consistency.

\subsection{Plan Module}
The Plan module orchestrates agent behavior in empathetic interactions.
To ensure that agents mimic human behavior in using stickers, we specially design three key actions to ensure the timely use of stickers, their coherence with the context, and their effective role in empathy.

\begin{itemize}[nolistsep, noitemsep]
    \item \textbf{\textit{Intention}}: 
    Directed by the Profile module, the agent assesses the use of emotive stickers, considering interaction history and observed data.
    \item \textbf{\textit{Query}}: 
    When intending to use a sticker, the agent describes its emotion, shaped by the Profile module, observations, and the intent for its use.
    \item \textit{\textbf{Select}}: 
    After querying its sticker database with the crafted query, the agent retrieves the top-K most relevant stickers. It organizes these stickers into a list detailing their emotion, description, and usage recommendation, from which it selects the most appropriate sticker.
\end{itemize}

\subsection{Action Module}
The Action module serves as the window and bridge for communication with the external world. It is responsible for receiving instructions and inputs from other parts of the system and executes two main actions based on this information:

\begin{itemize}[nolistsep, noitemsep]
\item \textit{\textbf{Message}}: 
The agent generates text responses based on the agent's Profile and Memory module.
\item \textit{\textbf{Sticker}}: 
The agent engages in sticker usage as dictated by the Plan module.
\end{itemize}

\subsection{Manager Agent}
The Manager Agent, a critical component in enhancing the performance of the \agent, plays a significant role in ensuring interaction quality and dialogue consistency. This agent comprises two main modules: 
\textbf{(1) Quality Assurance:}
This module plays a key role in maintaining high dialogue standards by evaluating sticker relevance to enrich conversations, regulating sticker usage to avoid misuse, and conducting quality checks focused on response length appropriateness and redundancy avoidance.
\textbf{(2) Process Control:}
This module takes charge of guiding the dialogue's progression. It leverages historical interactions as a reference to make informed decisions on the optimal timing for concluding dialogues. 

\subsection{The \dataset Dataset}
Leveraging \agent, we propose the construction of a multimodal empathetic dialogue dataset, \textbf{\dataset}.
\begin{figure*}[t]
    \centering
    \includegraphics[width=0.9\linewidth]{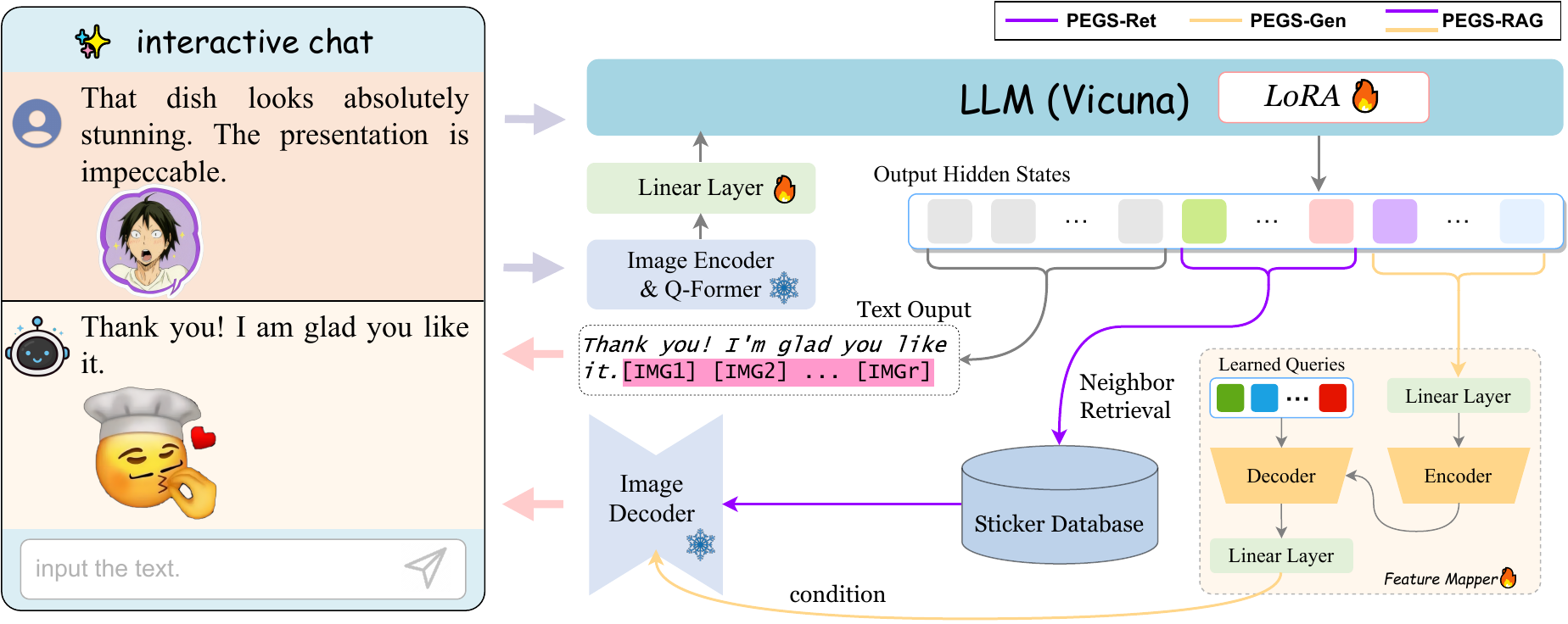}
    \caption{
    The architecture of \ourmodel framework includes various routing options, distinguished by colored connecting lines. Input stickers undergo joint encoding by an image encoder, Q-Former, and a linear layer, with Vicuna serving as the language model. The output of the LLM activates two sets of tokens differently across model versions: one for image retrieval and the other as a textual condition. Subsequently, the frozen image decoder generates images.}
    \label{fig:model_architecture}
    \vspace{-0.7em}
\end{figure*}
In our dataset creation, we define two roles: User, the dialogue initiator with a profile from 2,000 generated personalities, and System, acting as a listener and empathizer.
To mimic the human use of stickers, the sticker dataset is divided into 100 vector databases, each ensuring a consistent emotional distribution of stickers. The top-K of Tool module is set to 10.
For each dialogue session, both User and System access a randomly chosen vector database, mirroring human sticker usage and enriching sticker diversity. 

\dataset consists of 12,931 dialogue sessions, 67,505 stickers (unique 5.8K stickers) and 2K user personalities. 
Each session averages 5.22 stickers and 5.49 dialogue turns.  
Emotional label distribution analysis, as illustrated in Figure \ref{fig: compare_user_system}, highlights the differences in sticker usage between User and System, reflecting their unique roles.
Further analysis of the \dataset is detailed in the Appendix \ref{app: dataset_detail}.

\section{\ourmodel}
\label{sec:pegs}
We devise a multimodal empathetic response generation framework, \textbf{\ourmodel}, with the ability to \fourmodel. Figure \ref{fig:model_architecture} illustrates the architecture of our framework. With different strategies to generate images, we derive three models based on this framework: PEGS-Ret/Gen/RAG, which denote the retrieval, generation, and retrieval-augmented generation methods to provide images. Technically, we utilize ViT-G/14 from EVA-CLIP \citep{fang2023eva}, Q-Former from BLIP-2 \citep{li2023blip}, and a linear layer to encode images. Vicuna \citep{vicuna2023}, a widely used language model in LMMs, is employed for language modeling. Stable Diffusion (SD) \citep{rombach2022high} as the image decoder.

\subsection{Multimodal Input Perception}
With reference to existing works \citep{li2023blip, dai2023instructblip, zhu2023minigpt}, we convert the multimodal inputs into feature vectors that can be solved by LLM. Specially, each text token is embedded into a vector $e_t\in \mathbb{R}^{1\times d}$, while each image is first encoded by a pre-trained vision encoder, and then an aligned feature vector $e_v\in \mathbb{R}^{32\times d}$ is obtained via Q-Former and a linear projection layer.

\subsection{Multimodal Output Generation}

\paragraph{Expanding Vocabulary} We extend the vocabulary $V$ with an additional visual tokens set $V_\mathrm{img}=\{\mathrm{[IMG1]}, \mathrm{[IMG2]}, \ldots, \mathrm{[IMG\{} r \mathrm{\}]}\}$. We denote the original word embedding matrix as $E\in \mathbb{R}^{|V|\times d}$. For the embeddings matrix $E^*\in \mathbb{R}^{|V^*|\times d}$ of the extended vocabulary $V^* = V \cup V_{\mathrm{img}}$, the embeddings $E_{\mathrm{img}}\in \mathbb{R}^{r\times d}$ of the added special tokens are randomly initialized and the embeddings $E$ of the original token are preserved:

\vspace{-0.5em}
\begin{equation}
    \centering
    E^*[0: |V|, :] = E
    \label{vocab}
\end{equation}

We split visual tokens into two sets, where the front $k$ tokens for image retrieval and the back $r-k$ tokens for image generation:

\vspace{-1.5em}
\begin{align}
    \centering
    &V_{\mathrm{ret}} =\{\mathrm{[IMG1]}, \ldots, \mathrm{[IMG\{} k \mathrm{\}]}\} \\
    &V_{\mathrm{gen}}=\{\mathrm{[IMG\{}k+1\mathrm{\}]}, \ldots, \mathrm{[IMG\{} r \mathrm{\}]}\}
\end{align}

\noindent where $V_{\mathrm{ret}}$ is used in \pret and \prag and $V_{\mathrm{gen}}$ is used in \pgen and \prag.


\paragraph{Text Generation} Receiving multimodal inputs, the target is to generate joint sequences of text tokens and visual tokens $\{\mathrm{[IMG\{}i{\}]}\}_{i=1}^{r}$. Specifically, the generated token can be represented as $U=\{u_1, \ldots, u_k\}$, where $u_i\in V^*$. The loss function $\mathcal{L}_{\mathrm{lm}}$ is defined as:

\vspace{-1.5em}
\begin{equation}
\hspace{-0.6em}
    \centering
    \mathcal{L}_{\mathrm{lm}} \!=\! -\!\sum_{i=1}^k\! \log p(u_i|s, u_1, \ldots, u_{i-1}; \theta, E_{\mathrm{img}})
\end{equation}

\noindent where $s=\{e_{m}^{(1)}, e_{m}^{(2)}, \ldots, e_{m}^{(l)}\}$ and $m \in \{t,v\}$ denoting the modality.The original LLM weights $\theta$ are kept frozen, and we only update $E_{\mathrm{img}}$. 

\paragraph{Image Retrieval}
For image retrieval, \ourmodel aligns the hidden states $h_{\mathrm{ret}}$ corresponding to $V_{\mathrm{ret}}$ into the retrieval space by contrastive learning \citep{chopra2005learning}. 
${W}_{\mathrm{t}}\in\mathbb{R}^{d \times e} $ and ${W}_{\mathrm{i}} \in \mathbb{R}^{p \times e} $to bridge the semantic gap and adjust the dimension.
Cosine similarity is used to measure the similarity of the projection vectors:

\vspace{-1.5em}
\begin{equation}
    \centering
    \text{sim}(x, y) = \frac{({W}_{\mathrm{t}}^T h_{\mathrm{ret}}(x))^T({W}_{\mathrm{i}}^T \nu_\phi(y))}{\Vert{W}_{\mathrm{t}}^T h_{\mathrm{ret}}(x)\Vert \Vert{W}_{\mathrm{i}}^T \nu_\phi(y))\Vert}
\end{equation}

\noindent where $\nu_\phi$ is the image encoder. The projection vectors are used to minimize the InfoNCE loss \citep{oord2018representation}, which consists of text-to-image (t2i) and
image-to-text (i2t) loss in a batch of N text-image
pairs ($x_i$, $y_i$):

\vspace{-1.5em}
\begin{flalign}
    \hspace*{-0.73em}\mathcal{L}_{\mathrm{t2i}} \!&=\! -\frac{1}{N}\! \sum_{i=1}^N\!\left(\!\log\! \frac{\exp(\text{sim}(x_i, y_i)\!/\!\tau)}{\sum_{j=1}^N\! \exp(\text{sim}(x_i, y_j)\!/\!\tau)}\!\right)
    \\ \hspace*{-0.73em} \mathcal{L}_{\mathrm{i2t}} \!&=\! -\frac{1}{N} \!\sum_{i=1}^N\!\left(\!\log\! \frac{\exp(\text{sim}(x_i, y_i)\!/\!\tau)}{\sum_{j=1}^N\! \exp(\text{sim}(x_j, y_i)\!/\!\tau)}\!\right) 
\end{flalign}


\vspace{-0.5em}
\begin{equation}
    \centering
    \mathcal{L}_{\mathrm{ret}} = \frac{1}{2}(\mathcal{L}_{\mathrm{t2i}} + \mathcal{L}_{\mathrm{i2t}})
\end{equation}

$\mathcal{L}_{\mathrm{ret}}$ is the loss used to optimize the projection layers for retrieval.

\paragraph{Image Generation} We align the hidden states $h_{\mathrm{gen}}$ corresponding to the output visual tokens $V_{\mathrm{gen}}$ into the input space of the image decoder. Specifically, we connect them through a feature mapper module, containing two linear layers and a 4-layer encoder-decoder transformer model with a learnable queries feature $q$. For the given image caption $c$ (and its emotion $e$ if available), our target is to minimize the MSE loss between their embeddings derived from the frozen pre-trained SD text encoder $\eta$ and the projected representations:
\vspace{-0.5em}
\begin{equation}
    \centering
    \mathcal{L}_{\mathrm{gen}} = \lVert \theta_{\mathrm{Mapper}}(h_{\mathrm{gen}}, q) - \eta(c,\lbrack e \rbrack) \rVert_2^2
\end{equation}

\paragraph{Retrieval-Augmented Image Generation} 

Intuitively, continuing to do generation on retrieved images can extend the diversity of images while maintaining image quality, thus we explore retrieval-augmented generation.

Specifically, we retrieve an image serving as a latent representation $c_I$ for augmenting the generation process. During image generation, $h_{\mathrm{gen}}$ remains utilized as a condition.

\subsection{Joint Learning}
Pre-trained LLMs excel in text interactions yet struggle in empathetic conversations.
We further utilize the constructed \dataset for jointly fine-tuning of the entire model to achieve the capabilities of multimodal perception and generation. We train our model in an end-to-end manner, using LoRA \citep{hu2021lora} to synchronize the update of a limited number of parameters in the LLM with the input linear projection layers and the feature mappers. The overall loss function $\mathcal{L}$ can be represented as:

\vspace{-1em}
\begin{equation}
    \centering
    \mathcal{L} = 
    \lambda_1 \times \mathcal{L}_{\mathrm{lm}} + \lambda_2 \times \mathcal{L}_{\mathrm{gen}} + \lambda_3 \times \mathcal{L}_{\mathrm{ret}}
\label{eq: overall_loss}
\end{equation}

\noindent where $\lambda_1$, $\lambda_2$, and $\lambda_3$ represent hyperparameters. $\mathcal{L}_{\mathrm{gen}}$ includes emotion $e$ as an input. For \pret, $\mathcal{L}_\mathrm{gen}=0$, and for \pgen, $\mathcal{L}_\mathrm{ret}=0$. Implementation details can be found in Appendix \ref{app: pegs}.

\section{Evaluation Metrics}

\subsection{Text Metrics}
To comprehensively evaluate the fluency, diversity and accuracy of dialogue generation, we utilize an array of broadly recognized text metrics, comprising \textbf{BLEU} \cite{papineni-etal-2002-bleu}, Distinct-n (\textbf{Dist}-n) \cite{liDiversityPromotingObjectiveFunction2016}
, ROUGE-L (\textbf{ROU\_L}) \cite{kingmaAdamMethodStochastic2017}, METEOR (\textbf{MET}) \cite{banerjee-lavie-2005-meteor}, \textbf{CIDEr} \cite{vedantamCIDErConsensusBasedImage2015a}, BERTScore (\textbf{BERTS}) \cite{zhangBERTScoreEvaluatingText2020}.

\subsection{Multimodal Metrics}
We use MM-Relevance (\textbf{MMr}) \cite{feng-etal-2023-mmdialog} to assess the relevance between the predicted multimodal response and the golden response. 
However, MMr overlooks the frequency of multimodal replies, a critical aspect considering that text responses within a modality often show higher similarity than across modalities. This could result in models that rarely engage in multimodal responses receiving inaccurately high MMr scores. 
To address this issue, we impose penalties on models that are more inclined to generate text-only replies:

\vspace{-0.4em}
\begin{equation}
\label{eq: mmr}
f\text{-MMr.} = (1-\alpha(1-f))\cdot\text{MMr.}
\end{equation}
where $\alpha$ represents the penalty coefficient. In our experiments, we set $\alpha$ to 0.8. $f$ denotes the relative frequency at which the model produces multimodal responses, $f\in[0, 1]$. 

\begin{table*}[t]
  \centering
  \resizebox{0.85\textwidth}{!}{
    \begin{tabular}{ccccccc}
    \toprule
    \textbf{Model} & \textbf{BLEU-1/2/3/4} & \textbf{Dist-1/2/3} & \textbf{ROU\_L.} & \textbf{MET.} & \textbf{CIDEr} & \textbf{BERTS.} \\
    \midrule
    Vicuna-text & 0.44/0.30/0.22/0.17 & \textbf{0.879/0.994/0.999} & 0.31  & 0.37  & 0.39  & 0.878  \\
    Vicuna-tool & 0.43/0.29/0.22/0.17 & 0.870/0.989/0.994 & 0.30  & 0.36  & 0.38  & 0.900  \\
    \midrule
    ChatGLM3-text & 0.42/0.28/0.21/0.16 & 0.806/0.981/0.996 & 0.31  & 0.40  & 0.40  & 0.886  \\
    ChatGLM3-tool & 0.36/0.22/0.16/0.11 & 0.859/0.992/0.998 & 0.26  & 0.34  & 0.20  & 0.899  \\
    \midrule
    \pret & 0.46/0.32/0.25/0.20 & 0.839/0.989/0.997 & 0.34  & 0.42  & 0.47  & 0.906  \\
    \prag & 0.46/0.32/0.25/0.20 & 0.839/0.989/0.997 & 0.34  & 0.42  & 0.47  & 0.906  \\
    \pgen & \textbf{0.47/0.33/0.26/0.21} & 0.848/0.990/0.997 & \textbf{0.35} & \textbf{0.44} & \textbf{0.57} & \textbf{0.911} \\
    \bottomrule
    \end{tabular}%
    }
    \caption{Results of quality of text generate in \ourmodel and baseline models.}
    \vspace{-0.7em}
  \label{tab: text_result}%
\end{table*}%

\subsection{LLM-based Metrics}
LLMs are capable of grading similarly to humans, providing scores for both textual and sticker outputs, thus enabling a comprehensive multimodal evaluation system. 

We introduce three LLM-based metrics:
\textbf{(1) Empathy}: 
We assess empathy in model responses, both textual (Empathy-text, \textbf{EMP-txt}) and multimodal (Empathy-multimodal, \textbf{EMP-mm)}, averaging scores from five independent scorings to reduce randomness and bias.
\textbf{(2) Consistency}: 
Based on the context, we assign consistency scores for textual and sticker responses, marked as Consistency (\textbf{CON}), employing the same scoring method.
\textbf{(3) Rank}: 
We compare responses of different models to the same context. The ranker organizes responses based on quality, empathy, and consistency, averaging across many possible ranking combinations to ensure fairness and objectivity.


\subsection{Human Metrics}
Considering the subjectivity of empathy and the complexity of multimodal responses, we establish seven detailed manual evaluation metrics:
Sticker Generation Quality (\textbf{StiGQ}), 
Empathy Sticker (\textbf{Es}), 
Empathy Text (\textbf{Et}), 
Consistency (\textbf{Con}), 
Fluency (\textbf{Flu}), 
Informativity (\textbf{Inf}),
Sticker Diversity (\textbf{StiD}).

Metrics are scored on a 1 to 5 scale, with higher scores denoting better performance. For fairness and consistency, all models are evaluated in identical contexts. The evaluation panel consists of five members who score models anonymously, unaware if they are assessing a baseline model. The human evaluation process randomly selects 100 dialogue sessions from the \dataset test set.

\section{Experiments}
Experiments are conducted on \dataset. We perform response predictions for all turns of each dialogue and consider all previous turns as context.

\subsection{Implementation Details}
Our models are pre-trained on LAION115M, a sub-dataset of LAION400M \citep{schuhmann2021laion}, and fine-tuned on \dataset. More implementation details and hyperparameter settings are provided in Appendix \ref{app: pegs}. The training procedure is conducted on 2 NVIDIA A6000 48G GPUs.

\subsection{Baselines} 
We benchmark against Vicuna-7B \cite{vicuna2023} and ChatGLM3-6B \cite{zeng2023glm130b} models, employing two experimental paradigms: 

\vspace{-0.4em}
\paragraph{Text Fine-tuning} We fine-tune the models using text data from the \dataset training set and then make predictions on the test dataset.
\vspace{-0.4em}
\paragraph{Tool Learning} We fine-tune the models with multimodal data from the \dataset training set.
We use stable-diffusion as a tool to teach the models how to utilize it for generating sticker responses \cite{tangToolAlpacaGeneralizedTool2023}.
To ensure fairness, the tool models employ the same SD model as \ourmodel.

\subsection{Result Analysis}
Table \ref{tab: text_result} reports the results for text metrics.
\ourmodel's performance on the Dist-n is slightly lower than that of the Vicuna. This may be attributed to PEGS expanding Vicuna's vocabulary to facilitate sticker perception and generation, impacting PEGS's text diversity to a certain degree.
In our experiment, the training objective of tool learning includes both text generation and tool calling. These two objectives may interfere with each other. As a result, the tool model's text capabilities fall short of the text fine-tuning model's.
\ourmodel's end-to-end structure integrates multimodal inputs and outputs, streamlines tasks with unified training objectives, and yields the best text results.
\pgen utilizes fewer special tokens (32), thereby outperforming \pret and \prag in text metrics.
These findings corroborate the efficacy of \ourmodel framework in generating text responses of high quality and accuracy.

\begin{table}[htbp]
  \centering
  \scriptsize
  \resizebox{0.35\textwidth}{!}{
    \begin{tabular}{ccccc}
    \toprule
    \textbf{Model} & \textbf{Freq.} & \textbf{MMr.} & \textbf{f-MMr.} \\
    \midrule
    Vicuna-tool & 0.141 & \textbf{0.725} & 0.602   \\
    ChatGLM3-tool & 0.905 & 0.659  & 0.647  \\
    \midrule
    \pret & 0.850 & 0.674  & 0.653   \\
    \prag & 0.847 & 0.680  & \textbf{0.659}   \\
    \pgen & 0.811 & 0.672  & 0.647  \\
    \bottomrule
    \end{tabular}%
    }
    \caption{Results of multimodal metrics.}
    \vspace{-0.7em}
  \label{tab: multimodal_metrics}%
\end{table}%

Table \ref{tab: multimodal_metrics} displays the results for multimodal metrics, where \textbf{Freq} represents the relative frequency of stickers replies from each model. Although Vicuna-tool achieves a high MMr (72.48), this result is largely due to its lower Freq (0.141), which may not accurately represent its multimodal interaction capabilities.
F-MMr is more robust than MMr because it accounts for Freq.
\ourmodel excels in f-MMr, showcasing that its end-to-end structure, integrating text and stickers, achieves high consistency in multimodal reply generation.

\begin{table}[!htbp]
  \centering
  \small
  \resizebox{0.4\textwidth}{!}{
    \begin{tabular}{ccccc}
    \toprule
    \textbf{Model} & \textbf{EMP-txt} & \textbf{EMP-mm} & \textbf{CON.} & \textbf{Rank} \\
    \midrule
    Vicuna-text & 3.677  & -     & 4.322  & 4.527  \\
    Vicuna-tool & 3.822  & 3.799  & 4.247  & 4.093  \\
    \midrule
    ChatGLM3-text & 3.691  & -     & 4.341  & 5.037  \\
    ChatGLM3-tool & 3.700  & 3.760  & 4.220  & 4.400  \\
    \midrule
    \pret & 3.873  & 4.040  & 4.380  & 4.030  \\
    \prag & \textbf{3.927} & 4.076  & 4.370  & 3.900  \\
    \pgen & 3.768  & \textbf{4.353} & \textbf{4.404} & \textbf{1.917} \\
    \bottomrule
    \end{tabular}%
    }
    \caption{Results of LLM-based metrics.}
    \vspace{-0.7em}
  \label{tab: llm_score}%
\end{table}%
Table \ref{tab: llm_score} presents the evaluation results for the LLM. 
Relative to the baseline model, the two tool-learning models outperform the text model in ranking. This underscores the pivotal role of stickers in enhancing empathetic communication.
According to the results from LLM-based metrics, PEGS can produce empathetic texts while ensuring high consistency, and utilize stickers to enhance emotional resonance.
Table \ref{tab: multimodal_metrics} and \pgen's EMP-mm (4.353) indicates that the quality of the multimodal response directly correlates with the extent to which it enhances empathy.

\begin{table}[!htbp]
  \centering
  \small
  \resizebox{0.48\textwidth}{!}{
    \begin{tabular}{cccccccc}
    \toprule
    \textbf{Model} & \textbf{StiGQ.} & \textbf{Et.} & \textbf{Es.} & \textbf{Con.} & \textbf{Flu.} & \textbf{Inf.} & \textbf{StiD.} \\
    \midrule
    Vicuna-tool & 4.09  & 4.07  & 3.78  & 4.08  & 4.23  & 4.08  & 3.20  \\
    ChatGLM3-tool & 4.32  & 4.06  & 3.24  & 4.11  & \textbf{4.58} & 3.99  & 3.10  \\
    \midrule
    \pret & -     & 4.11  & 4.17  & 4.22  & 4.36  & 4.09  & 3.40  \\
    \prag & 3.37  & 4.12  & 2.92  & 4.22  & 4.36  & 4.10  & \textbf{3.80} \\
    \pgen & \textbf{4.53} & \textbf{4.29} & \textbf{4.19} & \textbf{4.37} & 4.47  & \textbf{4.26 } & 3.60  \\
    \bottomrule
    \end{tabular}%
    }
    \caption{Results of the human evaluation.}
    \vspace{-0.7em}
  \label{tab: human_evaluation}%
\end{table}%

Table \ref{tab: human_evaluation} shows the result of human evaluation.
\ourmodel notably outstrips the two baseline models across most evaluative indicators, save for Flu, thus corroborating the efficacy of the \ourmodel framework.
The Flu of \ourmodel ($<$4.5) is lower compared to ChatGLM3-tool (4.58), likely because its high Inf ($>$4.09) results in a slight reduction in Flu.
According to Inf and Et, the amount of information is a crucial factor influencing empathy.
\prag excels in sticker diversity (3.8) but records the lowest score in sticker generation quality (3.37). This observation suggests that although the RAG strategy enhances sticker variety, it might also reduce the quality of sticker generation in this experiment. 
Analysis of StiGQ, Es, and StiD reveals that multimodal responses can enhance empathy, and this enhancement is positively correlated with the responses' quality.
When considering all evaluation of human metrics, \pgen stands out remarkably, partly because of its minimal use of special tokens and partly due to its proficient sticker generation capability (owe to end-to-end structure). These outcomes highlight \pgen's exceptional ability to produce high-quality, emotionally resonant, and diverse multimodal responses.

Further analysis is presented in Appendix \ref{app: further_result}. We collect the case study in Appendix \ref{app: case_study}.

\section{Conclusion}
We explored the concept of multimodal empathetic response and created the \dataset, the first dataset specifically designed for multimodal empathetic dialogue. We developed the \fagent, a sophisticated LLM-based multi-agent system capable of simulating human-like interactions using stickers, thereby creating multimodal empathetic responses. Building upon the \dataset, we developed \fourmodel (\ourmodel), an advanced multimodal empathetic dialogue framework. This framework adeptly perceives and generates stickers, effectively enhancing the communication experience. Furthermore, we established a comprehensive set of evaluation metrics for multimodal empathy tasks based on LLM. We are confident this work will be a valuable asset in advancing research in the field of multimodal empathetic dialogue systems.

\section*{Limitations}
Although our study has achieved certain advancements, it is not without its limitations. Firstly, \dataset is exclusively in English, implying that our model and evaluations might not be directly translatable to other linguistic contexts. 
Secondly, although we investigate the retrieval-augmented generation (RAG) variant of the \ourmodel, hoping to enhance the quality and diversity of sticker generation, the RAG version has demonstrated instability in generating quality stickers in practice. Despite these issues, the pursuit of diversity and quality in sticker generation means that the explorations into the RAG version remain valuable. It suggests a promising research trajectory: how to elevate the stability and variety of the generated content by refining algorithms or amalgamating additional technological solutions.
Finally, from our experiments, \ourmodel and two baseline tool-model exhibit abnormal frequencies in the usage of stickers. Aligning the model's sticker usage frequency and timing with human-like interaction patterns is an important area for future exploration.

\section*{Ethics Statement}
The \dataset dataset, produced by the large language model (LLM), therefore sidesteps privacy concerns. Nonetheless, LLMs may inherently harbor biases, and both \dataset and \ourmodel could inadvertently magnify these biases in their application. 
Despite our efforts to mitigate these biases by incorporating a diverse range of personality information into \agent, it is important to acknowledge that biases intrinsic to LLMs cannot be completely eradicated. This inherent bias is an ongoing challenge that needs to be addressed continuously through iterative improvement and monitoring of the model.
PEGS reveals its robust capability in creating multimodal empathetic replies, adept at grasping human emotions and generating pertinent multimodal content. Such abilities, if exploited, could potentially manipulate user emotions. To forestall these risks, we have implemented a mechanism for detecting NSFW content in the generation of multimodal content. It is imperative to underscore that, despite our model's proficiency in aiding the understanding and response to human emotions, it is not intended to entirely supplant human emotional communication. Excessive dependency on AI for empathetic interactions might erode human empathic abilities and the cultivation of emotional intelligence.


\bibliography{acl,custom,zyq}

\clearpage

\appendix

\section{Dataset Statistics}
\label{app: dataset_detail}
In our dataset creation, it is important to note that vector databases numbered 1 to 80 are exclusively employed for the creation of training and cross-validation sets, whereas those numbered 81 to 100 are reserved for the construction of the test set. We configure the agent's "\textbf{\textit{select}}" action's top-K parameter to 10.

Table \ref{table:dataset_statistics} offers a statistical breakdown of the dataset, segmented into training, validation, and test splits. It provides a quantitative overview, detailing the number of instances in each subset, the unique stickers, turn counts, and token counts.
Our dataset is comparable in size to PhotoChat \cite{zang2021photochat}. To our knowledge, it is the first multimodal empathetic dialogue dataset, uniquely integrating stickers as non-textual elements to enhance empathetic communication more effectively.

\begin{table}[!htbp]
  \centering
  \resizebox{0.45\textwidth}{!}{
    \begin{tabular}{cccccc}
    \toprule
    split & Number & Unique Sticker & Turn  & $\text{Token}_\text{vicuna}$ & $\text{Token}_\text{GPT}$ \\
    \midrule
    train & 10,785 & 4,798  & 59,424 & 10,681,108 & 9,070,221 \\
    validation & 1,000  & 880   & 5,496  & 997,569 & 845,759 \\
    test  & 1,146  & 1,439  & 6,128  & 659,695 & 773,473 \\
    \bottomrule
    \end{tabular}%
}
  \caption{The statistics of \dataset.}
  \vspace{-0.5em}
  \label{table:dataset_statistics}%
\end{table}

\begin{figure}[!htbp]
    \centering
    \includegraphics[width=1.0\linewidth]{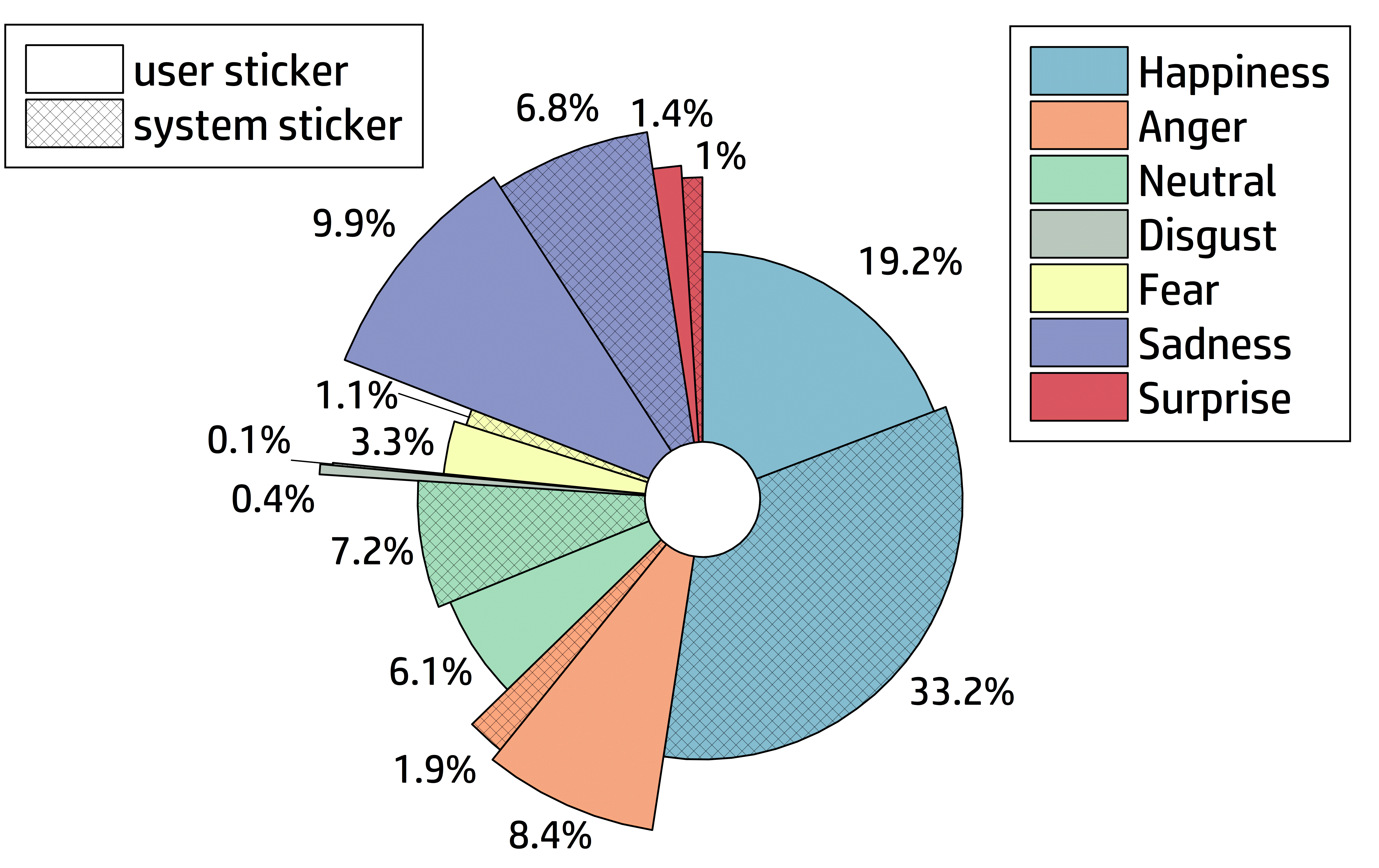}
    \caption{The chart of emotional distribution in the choice of stickers between the User and the System.}
    \label{fig: compare_user_system}
\end{figure}

\begin{figure*}[!htbp]
    \centering
    \includegraphics[width=1.0\linewidth]{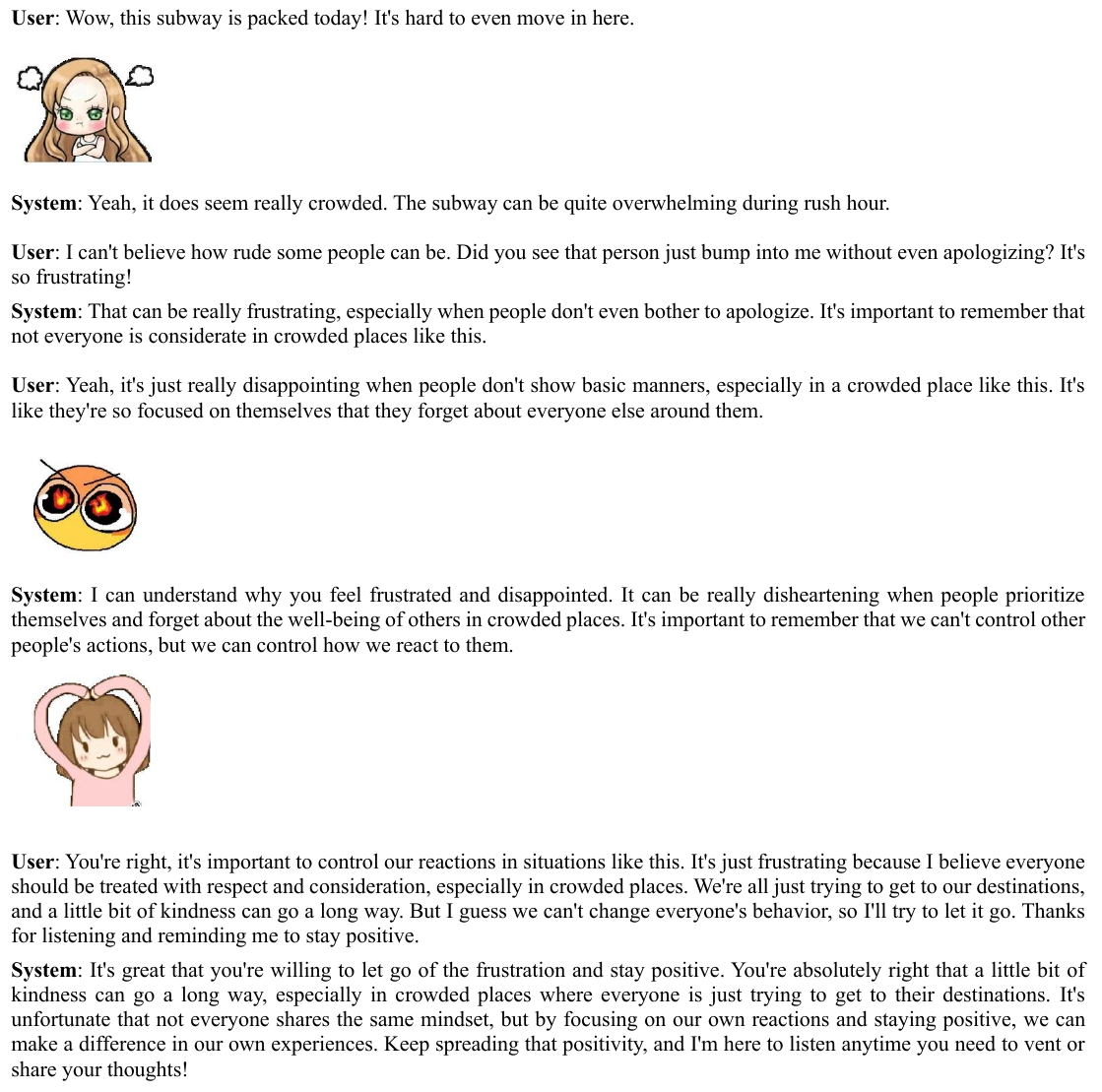}
    \caption{Example I from \dataset.}
    \label{fig: dataset_example_app1}
\end{figure*}

\begin{figure*}[!htbp]
    \centering
    \includegraphics[width=0.9\linewidth]{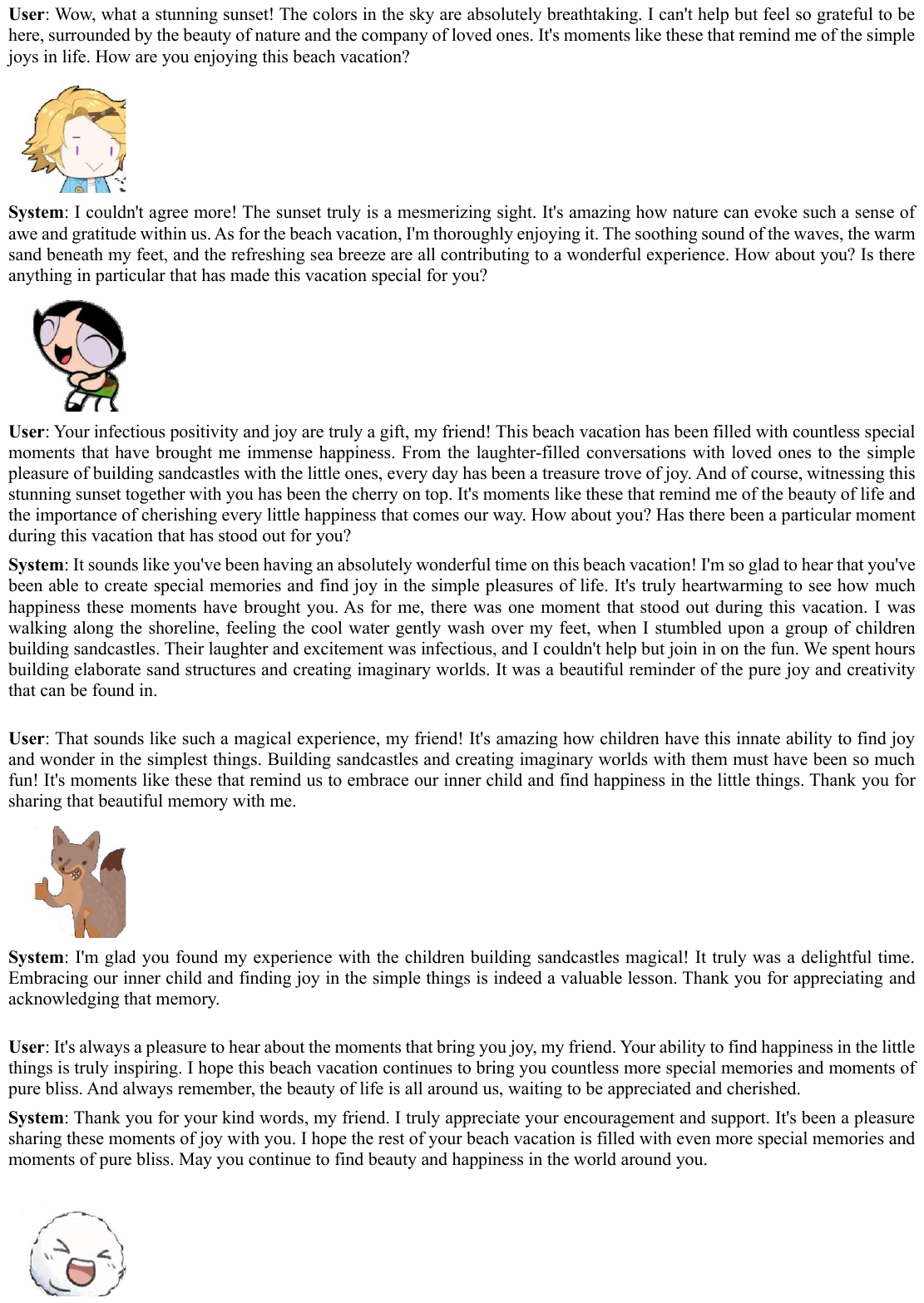}
    \caption{Example II from \dataset.}
    \label{fig: dataset_example_app2}
\end{figure*}

Figure \ref{fig: compare_user_system} shows the emotional distribution of stickers used by the User and the System. It reveals a striking trend: users have a significant preference for stickers that convey negative emotion, in contrast to the system. The system predominantly utilizes stickers to express neutral and positive emotion. This comparison not only reflects the distinct emotional expression preferences between users and the system but also highlights the system's active and supportive role in interactions.
Figure \ref{fig: dataset_example_app1} and \ref{fig: dataset_example_app2} present two examples from \dataset.

\begin{figure}[!htbp]
    \centering
    \includegraphics[width=1.0\linewidth]{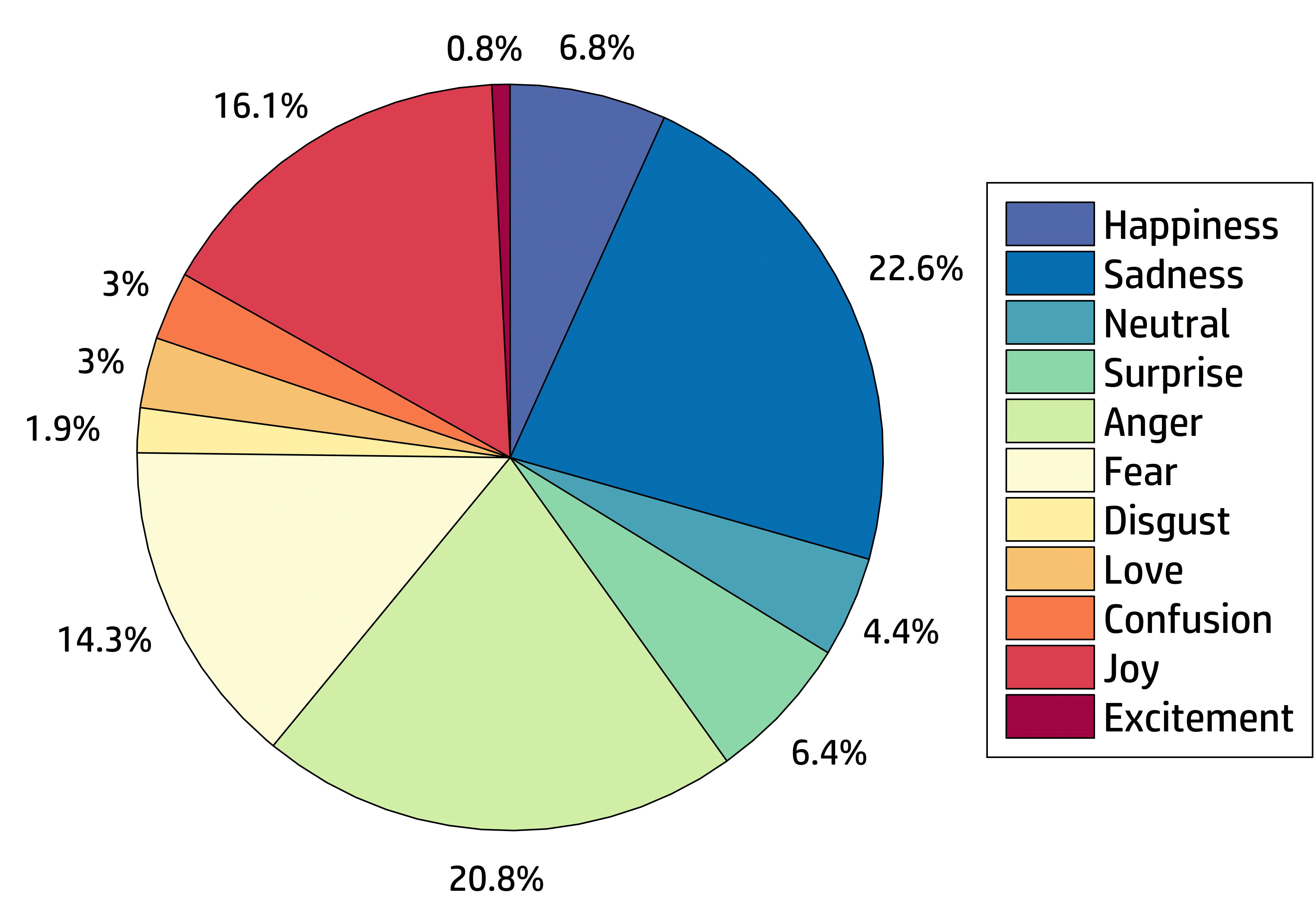}
    \caption{Emotion distribution of user profile in \fagent.}
    \label{fig:profile distribution}
\end{figure}

\begin{figure*}[!htbp]
    \centering
    \includegraphics[width=0.7\linewidth]{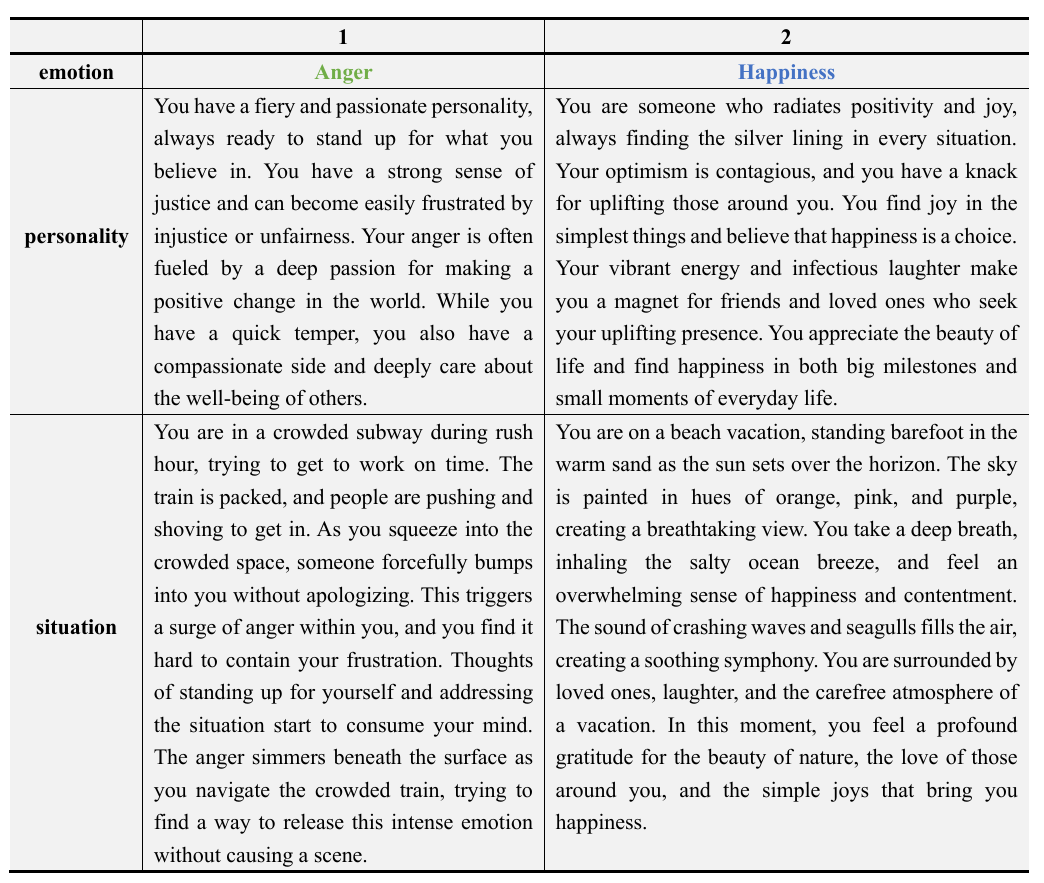}
    \caption{Example Profile of \agent.}
    \label{fig: profile_example}
\end{figure*}

Figure \ref{fig:profile distribution} showcases the emotion distribution within user profiles in the \fagent. The chart illustrates the percentage share of each emotion, offering insight into the prevalence of affective expressions within the dataset. Increasing personality distributions enhances the \agent's diversity, thereby helping to mitigate the inherent bias issue in LLMs.
Figure \ref{fig: profile_example} shows a profile example across various emotions.

\begin{figure}[!htbp]
    \centering
    \includegraphics[width=1.0\linewidth]{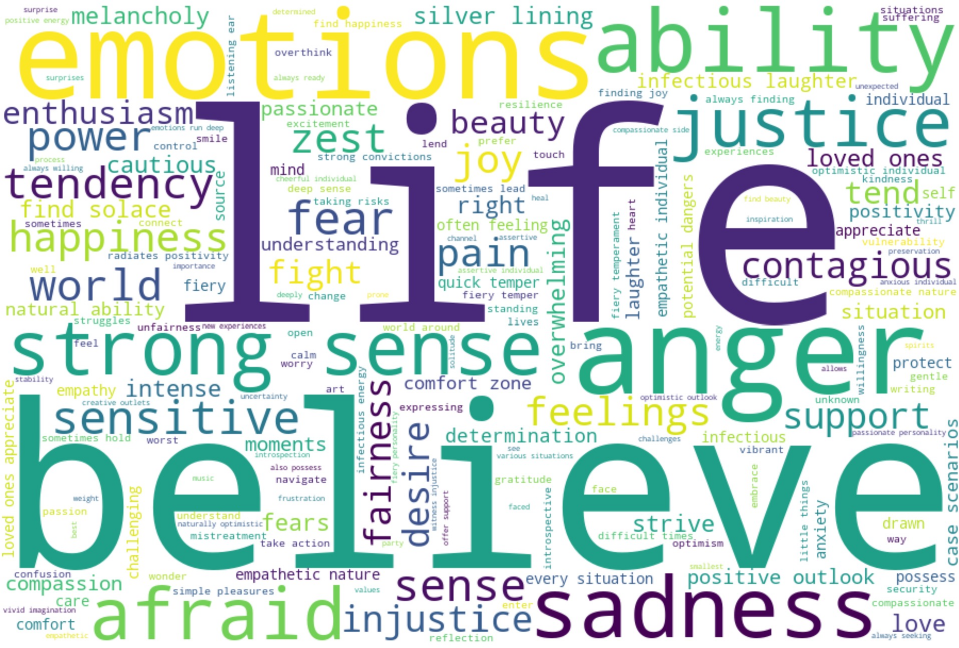}
    \caption{The 200 most popular emotion-related words in \dataset.}
    \label{fig:wordcloud}
\end{figure}

Figure \ref{fig:wordcloud}, a word cloud, visualizes the 200 most prevalent emotion-related words extracted from the \dataset. The prominence of each term in the cloud is indicative of its frequency within the dataset. This figure serves to underscore the linguistic diversity and the emotional range encapsulated in the dataset.

\section{Implementation Details of \agent}
\subsection{Tool Module}
\label{app: tool_module}
Before advancing to the follwing process, we altered the background of the SER30K stickers from black to white. The detailed process of building the \textbf{Tool module} is as follows:
\paragraph{Filter} 
Leveraging the LLaVa-v1.5-13b \citep{liuLlavav1ImprovedBaselines2023}, we craft a nuanced Chain of Thought (CoT) \cite{weiChainofThoughtPromptingElicits2023} procedure to meticulously scrutinize each SER30K sticker. This approach entails the model initially summarizing the sticker's content, subsequently assessing its relevance to casual conversation, and finally determining suitability through iterative responses. Each sticker undergoes multiple evaluations to ensure accuracy. 

\paragraph{Annotation} 
Filtered stickers undergo a second CoT analysis to answer 5 questions about their content, emotions, usage contexts, tone, and any comedic or satirical elements. This detailed annotation process provides deep insights, structured as Q\&A pairs for further use.

\paragraph{Knowledge} 
Leveraging ChatGPT-3.5-Turbo, we extract critical information from the Q\&A pairs, generating a comprehensive understanding of each sticker, including its description, emotion, and usage recommendation.

\paragraph{Split} 
We segment the filtered dataset into smaller vector databases, incorporating the original emotional labels alongside freshly acquired knowledge. Each database maintains an emotional sticker distribution that aligns with that of the original dataset.

Following the process described, we obtain a sticker database containing 10,648 stickers. 
The metadata for each sticker includes the original emotion label, the description of emotion, the description of content, and usage recommendations.
The sticker database is then evenly distributed into 100 smaller vector databases based on emotional balance.

\subsection{Manager Agent}
During dataset creation, we limit conversation turns to a maximum of six. The manager agent is programmed to review the conversation every two turns, and to verify the consistency of each sticker used with the conversation context. We observe that with more conversation turns, the likelihood of the agent repeating previous statements increased. Upon detecting such repetition, the manager agent will transition the agent's LLM to GPT-4 to prevent further repetition.

\begin{table*}[!htbp]
  \centering
  \resizebox{0.9\textwidth}{!}{
    \begin{tabular}{cccccccc}
    \toprule
    & \textbf{Model} & \textbf{learning rate} & \textbf{warmup steps} & \textbf{weight decay} & \textbf{batch size} & \textbf{max length} & \textbf{steps} \\
    \midrule
    Input & PEGS & 1e-4 & 2,000 & 0.05  & 128  & 32  & 80,000  \\
    \midrule
    \multirow{3}{*}{Output} & \pret & 3e-5 & 4,000 & 1e-4  & 36  & 77  & 142,220  \\
    \multirow{3}{*}{} & \prag & 3e-5 & 4,000 & 1e-4  & 36  & 77  & 142,220  \\
    \multirow{3}{*}{} & \pgen & 1e-5 & 1,000 & 0.05  & 108  & 32  & 80,000  \\
    \bottomrule
    \end{tabular}%
}
    \caption{Hyperparameters for pre-training PEGS.}
  \label{table: hyperparameters_pretraining}%
\end{table*}%

\begin{table*}[!htbp]
  \centering
  \resizebox{0.8\textwidth}{!}{
    \begin{tabular}{cccccccc}
    \toprule
    \textbf{Model} & \textbf{learning rate} & \textbf{warmup steps} & \textbf{weight decay} & \textbf{batch size} & \textbf{max length} & \textbf{epochs} \\
    \midrule
    \pret & 3e-5 & 70 & 0.05  & 36  & 77  & 10  \\
    \prag & 3e-5 & 70 & 0.05  & 36  & 77  & 10  \\
    \pgen & 5e-5 & 200 & 0.05  & 108  & 32  & 4  \\
    \bottomrule
    \end{tabular}%
}
    \caption{Hyperparameters for fine-tuning the output side of the PEGS on sticker-text pairs.}
  \label{table: hyperparameters_finetuning}%
\end{table*}%

\begin{table*}[!htbp]
  \centering
  \resizebox{\textwidth}{!}{
    \begin{tabular}{cccccccc}
    \toprule
    \textbf{Model} & \textbf{learning rate} & \textbf{warmup steps} & \textbf{weight decay} & \textbf{LoRA-r/alpha/dropout} & \textbf{batch size} & \textbf{max length} & \textbf{epochs} \\
    \midrule
    Vicuna-text & 5e-4 & 300 & 0.00 & 8/16/0.05  & 16  & 4,096  & 5  \\
    Vicuna-tool & 5e-4 & 300 & 0.00 & 8/16/0.05  & 16  & 4,096  & 5  \\
    \midrule
    ChatGLM3-text & 1e-4 & 300 & 0.10 & 8/16/0.05  & 16  & 4,096  & 5  \\
    ChatGLM3-tool & 1e-4 & 300 & 0.10 & 8/16/0.05  & 16  & 4,096  & 5  \\
    \midrule
    \pret & 3e-6 & 2,000 & 0.05 & 8/16/0.05  & 4  & 768  & 9  \\
    \prag & 3e-6 & 2,000 & 0.05 & 8/16/0.05  & 4  & 768  & 9  \\
    \pgen & 1e-4 & 1,000 & 0.05 & 8/16/0.05  & 4  & 768  & 4  \\
    \bottomrule
    \end{tabular}%
}
    \caption{Hyperparameters for fine-tuning baselines and PEGS on \dataset.}
  \label{table: hyperparameters_jointlearning}%
\end{table*}%
\section{More Implementation Details of PEGS}
\label{app: pegs}

\subsection{Pre-training}

\paragraph{Multimodal Perception} Our effort is built on BLIP-2 \citep{li2023blip}, which has captured vision-language knowledge from numerous aligned image-text pairs. \citet{zhu2023minigpt} have proven the effectiveness that freezing the pre-trained vision encoder, Q-Former, and LLM, while only pre-training a linear projection layer. We train the input linear projection layer using LAION115M.

\paragraph{Multimodal Generation} Distinguished from realistic images, stickers have an abstract nature and rich emotional expressions. Generic text-to-image models perform well in various scenarios, but they are flawed in sticker generation. Constrained by limited sticker data, there are not enough image-text pairs to turn the base text-to-image model to the style of stickers, and to pre-train for multimodal generation. As a remedy, we adopt a variant of CuteYukiMix\footnote{\url{https://civitai.com/models/28169?modelVersionId=163923}}, a SD model trained on cute cartoon data, as the pre-trained text-to-image model. In the pre-training phase, our objective is to align the representations with the text embeddings, so we train our model on the captions from LAION115M.

The AdamW optimizer is adopted, and the cosine annealing scheduler is used to adjust the learning rate. Table \ref{table: hyperparameters_pretraining} shows the hyperparameter settings in the pre-training phase.

\subsection{Fine-tuning}
\paragraph{Multimodal Generation} Pre-trained diffusion models (focusing on obvious features, such as characters' bodies, costumes) are insensitive to sentiment conditions, and it is difficult to directly perform end-to-end fine-tuning to learn emotion elements, we inserted a supervise fine-tuning (SFT) procedure on the image-text pairs with emotion to efficiently cause them to focus on emotion elements. Table \ref{table: hyperparameters_finetuning} shows the hyperparameter settings for this phase.

\paragraph{Joint Learning} We utilize the constructed \dataset for fine-tuning of the entire model to achieve the capabilities of multimodal perception and generation. We train our models in an end-to-end manner. The AdamW optimizer is adopted, and the cosine annealing scheduler is used to adjust the learning rate. Table \ref{table: hyperparameters_jointlearning} shows the hyperparameter settings in the joint-learning phase. In this work, $\lambda_1=1$, $\lambda_2=1$, $\lambda_3=1$ in Eq \ref{eq: overall_loss}.

\section{Further Result Analysis}
Table \ref{table: hyperparameters_jointlearning} presents the hyperparameters for fine-tuning baselines and \ourmodel. 
The end-to-end architecture of PEGS necessitates significant graphics memory during training, limiting it to smaller batchsize (4) and maximum lengths (768). Conversely, the baseline model accommodates larger batchsize (16) and maximum lengths (4096). 
These parameters suggest that \ourmodel can be further optimized for use on devices with larger graphics memory.

\label{app: further_result}
\subsection{Analysis of Multimodal Metrics}
From Table \ref{tab: multimodal_metrics}, taking into account the relative frequency of multimodal responses, f-MMr proves to be more reasonable and robust compared to MMr.
However, the multimodal response frequency of all models appears to be unusual, with the Vicuna-tool's frequency being notably low (0.141) and that of other models excessively high ($>$0.8). 
For comparison, the System's sticker usage frequency in the \dataset is 0.6.
This outcome suggests that future research should focus on adjusting the multimodal response frequency and timing of models to better suit human interaction.

In this paper, we do not use traditional image generation evaluation metrics, such as FID \citep{heusel2017gans}, IS \citep{salimans2016improved}.
These metrics are applied to evaluate the quality and diversity of realistic images. FID and IS are obtained by pre-trained Inception v3 \citep{szegedy2016rethinking} (pre-trained on realistic images) after certain processing. However, sticker possesses an abstract style and characteristics that do not match the application domains of FID and IS. Based on the above facts, we evaluate the quality and diversity of image generation by human evaluation.

\subsection{Analysis of LLM-based Metrics}
When using an LLM for scoring, to eliminate randomness, we repeat the scoring for each metric 5 times and calculate the average to determine the final score.
From Table \ref{tab: llm_score}, the EMP-mm of \ourmodel outperforms the two tool models, likely owing to \ourmodel's end-to-end framework that ensures greater consistency between text and stickers.
The top-ranking performance of \pgen further demonstrates the positive impact of text-sticker relevance on empathy.

\subsection{Analysis of Human Metrics}
\label{app: further_human_result}
\begin{figure*}[!htbp]
    \centering
    \includegraphics[width=\linewidth]{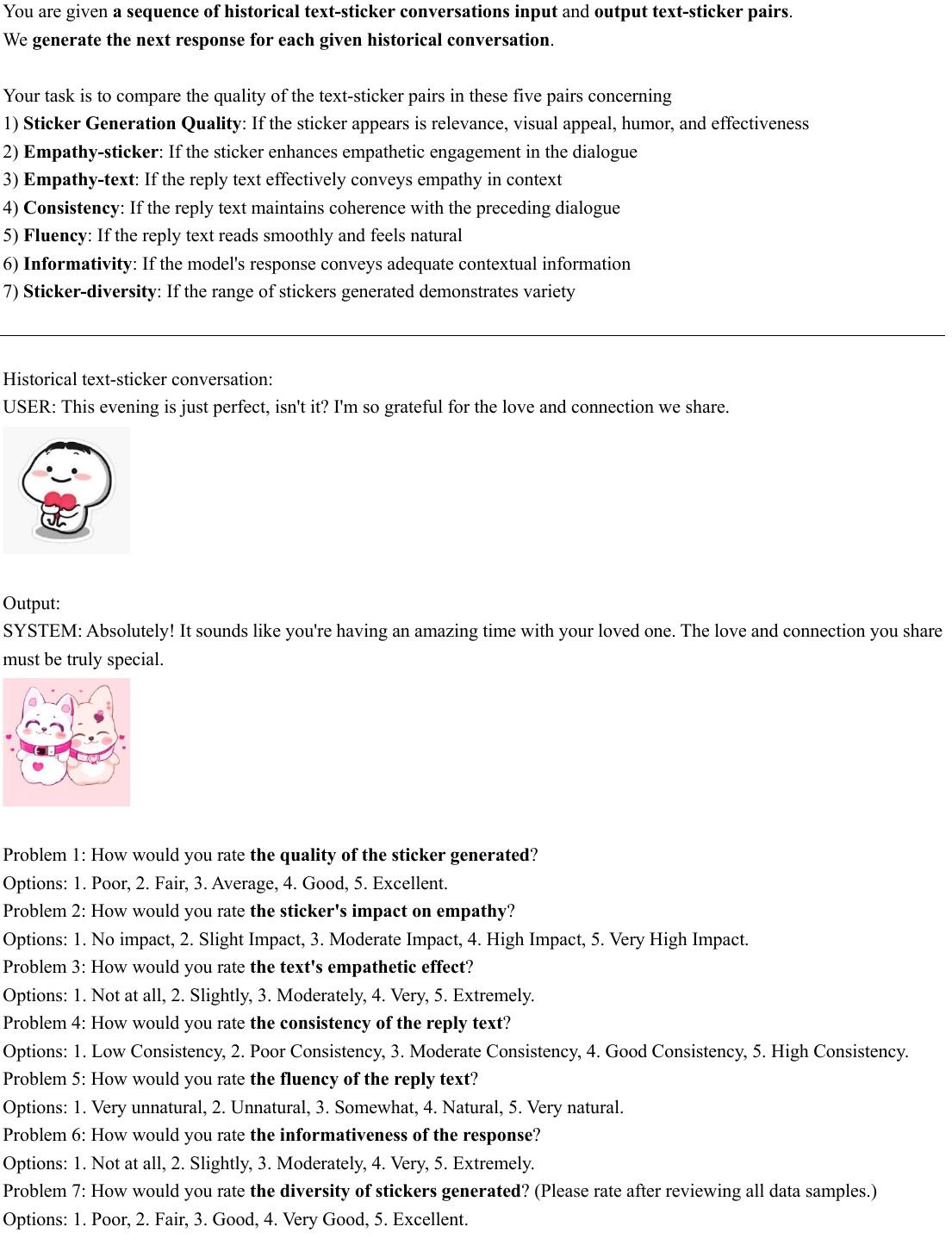}
    \caption{Human Evaluation Questionnaire.}
    \label{fig: human_eval_template}
\end{figure*}

Figure \ref{fig: human_eval_template} shows the complete questionnaire of human evaluation, which includes the detailed description and scoring criteria for each metric. The five scorers are graduate students in computer science, specializing in research on emotional reasoning or empathetic dialogue, with a thorough understanding of empathy tasks. During scoring, they are unaware of which model serves as the baseline and which is the \ourmodel.
\begin{table}[!htbp]
  \centering
  \resizebox{0.48\textwidth}{!}{
    \begin{tabular}{cccccccc}
    \toprule
    \textbf{Metric} & \textbf{StiGQ.} & \textbf{Et.} & \textbf{Es.} & \textbf{Con.} & \textbf{Flu.} & \textbf{Inf.} & \textbf{StiD.} \\
    \midrule
    ICC   & 0.920  & 0.540  & 0.928  & 0.457  & 0.728  & 0.608  & 0.365  \\
    \bottomrule
    \end{tabular}%
    }
    \caption{ICC Analysis Results for Human Evaluators}
  \label{tab: ICC}%
\end{table}%

Table \ref{tab: ICC} shows the results of the inter-rater reliability analysis conducted on the five scorers involved in the manual evaluation, using the Intraclass Correlation Coefficient (ICC) \cite{weirQUANTIFYINGTESTRETESTRELIABILITY2005a} as the assessment criterion.
Human scorers demonstrate high consistency in metrics like StiGQ, Es, Flu and Inf. 
The ICC of Et scores are moderate, likely because assessing empathy through text is highly subjective and depends significantly on individual life experiences.
Similarly, the moderate ICC scores for Con may stem from the nature of our task as a multimodal empathetic response task. Both the context and response involve multimodal information (sticker), and the text is affected by the stickers. This can, to some extent, disrupt the scorers' assessment of Con.
The ICC of StiD is poor ($<$0.4), primarily due to it being a global metric with a single score per model, leading to higher volatility.
\begin{table}[htbp]
  \centering
  \resizebox{0.48\textwidth}{!}{
    \begin{tabular}{cccccccc}
    \toprule
    \textbf{Metrics} & \textbf{EMP-txt} & \textbf{EMP-mm} & \textbf{CON.} & \textbf{Rank} & \textbf{ID.} & \textbf{f-MMr.} \\
    \midrule
    Spearman  & 0.40  & 0.70  & 0.87  & 0.40  & 0.80  & 0.90  \\
    Kendall & 0.40  & 0.60  & 0.74  & 0.40  & 0.60 & 0.80  \\
    \bottomrule
    \end{tabular}%
    }
    \caption{Correlation between the metrics and human annotation.}
  \label{tab: correlation}%
\end{table}%

Table \ref{tab: correlation} illustrates the correlation between various evaluative metrics and human annotations, as evidenced by Spearman's and Kendall's correlation coefficients. 
The EMP-txt metric calculates the correlation with the Et of human metrics, whereas EMP-mm corresponds to the Es. CON is matched with the Con, and Rank correlation is determined by averaging all scores from human evaluations. The ID metric measures correlation with the StiD of human metrics, and f-MMr matches the average of Con, Flu, Inf, and StiD. Both CON and f-MMr exhibit the highest consistencies under two coefficients, achieving 0.87 and 0.90 for Spearman and 0.74 and 0.80 for Kendall, respectively.

These results emphasize the precision and uniformity of these metrics in mirroring human assessments. 

EMP-mm and ID metrics also exhibit substantial correlations, suggesting a significant alignment with human perceptions within these domains.
Calculating the correlation between the results of LLM-based metrics and the results of human metrics demonstrates that with carefully designed prompts, LLM can effectively evaluate empathy tasks.

In summary, our experiments show that \ourmodel excels in producing varied and richly informative empathetic text responses, ensuring high contextual consistency, and generating diverse, high-quality stickers to enhance empathetic effects. These findings highlight the benefits of PEGS's end-to-end architecture.
The success of the \ourmodel demonstrates the effectiveness of \agent and \dataset. Considering the cost of \agent (shown in Appendix \ref{app: costing}), once \ourmodel training is completed, inference can be deployed locally to balance cost-effectiveness and efficiency.
This paper establishes a valuable resource for future research on multimodal empathetic response generation.

\section{Case Study}
\label{app: case_study}
\begin{figure*}[t]
    \centering
    \includegraphics[width=\linewidth]{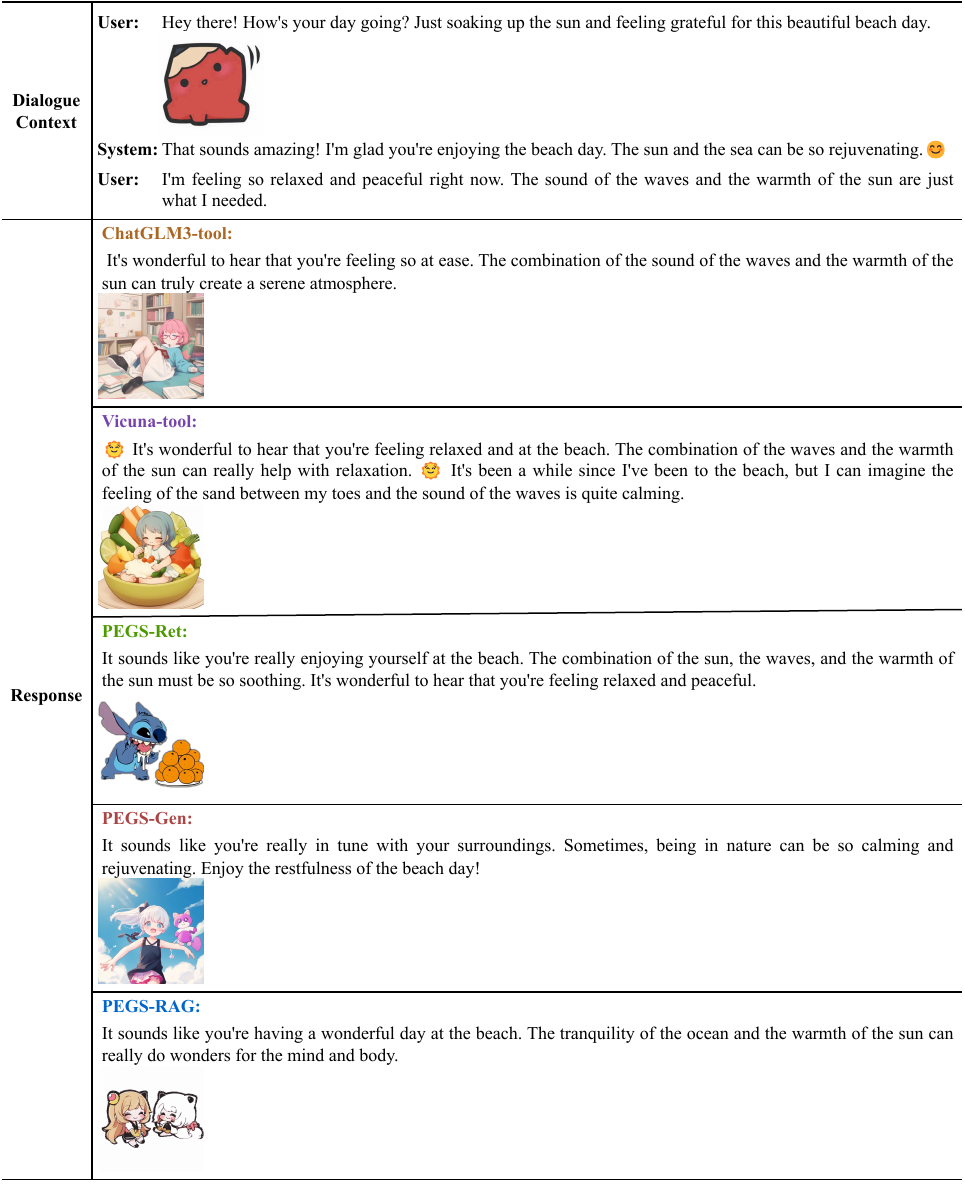}
    \caption{An example of \dataset test set. Top: the multimodal dialogue context between “User” and “System”. Bottom:
the multimodal responses generated by our designed multimodal baselines and models.}
    \label{fig:case_study}
\end{figure*}

To further investigate the quality of the multimodal responses predicted by our proposed framework, we show an example of \dataset test data in Figure \ref{fig:case_study}. As we have seen, the textual responses generated by multimodal baselines and our models are consistent with the dialogue context in terms of content and emotion, demonstrating the effectiveness of fine-tuning based on \dataset. All models provide visual responses (stickers) within the given context. The difference is that PEGS has more pronounced emotional tendencies, which facilitates empathy.

\begin{figure*}[t]
    \centering
    \includegraphics[width=0.8\linewidth]{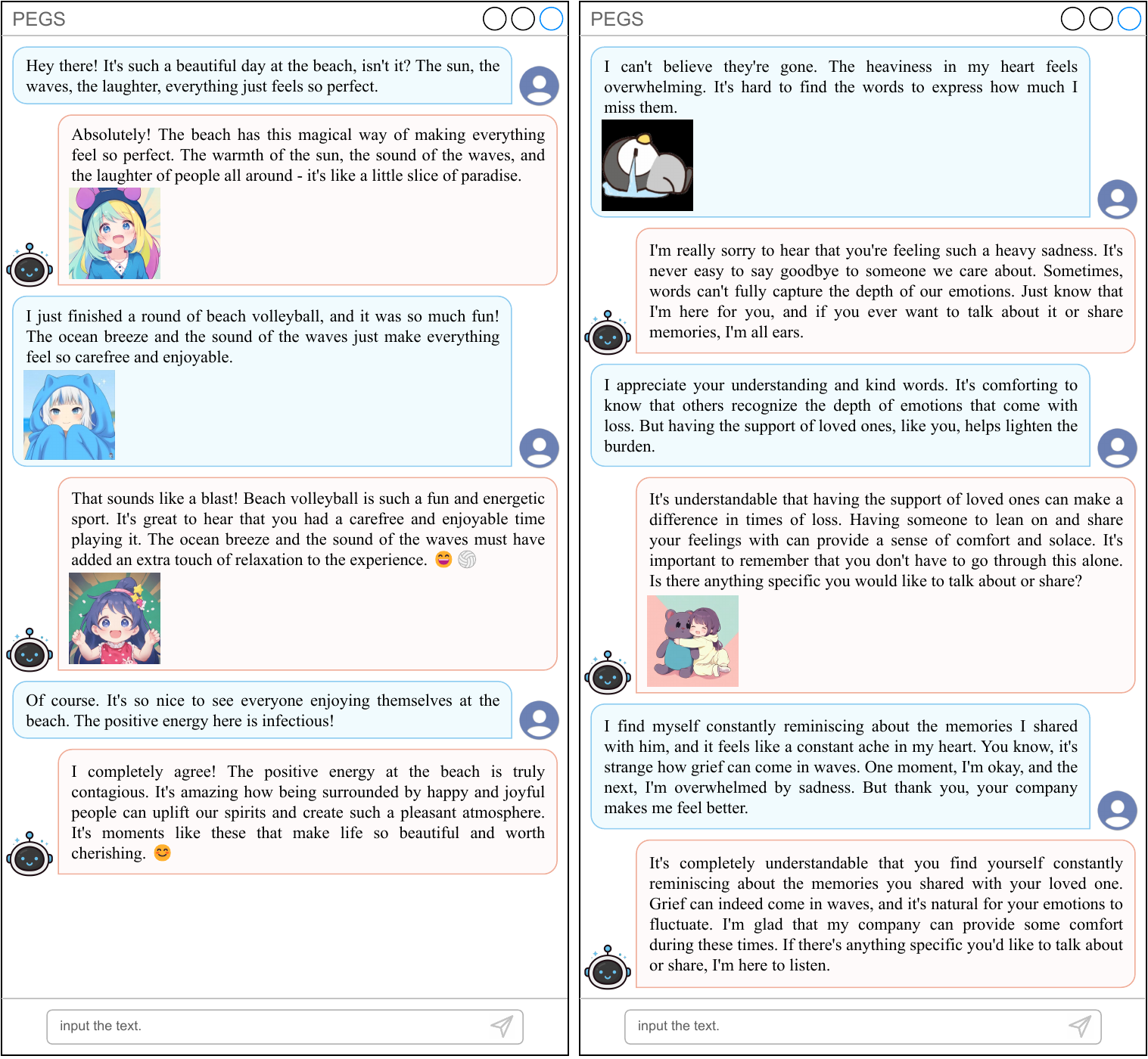}
    \caption{Examples of conversations by users interacting with PEGS. Users can chat with multimodal content (text and stickers) and will receive multimodal empathetic responses. Left: a conversation characterized by positive emotion (happiness). Right: a conversation characterized by negative emotion (sadness).}
    \label{fig:case_study_conversation}
\end{figure*}

To illustrate more straightforwardly the performance of our model in multimodal empathic conversations, we show several cases of user-chatbot interaction. As shown in Figure \ref{fig:case_study_conversation}, these cases demonstrate the ability of \ourmodel to empathize with users and perceive the stickers provided by users, along with generating multimodal responses with vivid stickers. For the positive dialogue context, our model generates positive stickers to enhance interactivity and emotional expression. For the negative dialogue context, our model reassures the user by an appropriate sticker (the hugging behavior in Figure \ref{fig:case_study_conversation} right).

\section{Prompts}
All \textcolor[RGB]{0,0,204}{\rm{\{format\_instructions\}}} in the prompt utilize the response of langchain\footnote{\url{https://www.langchain.com/}}'s response\_schema function implements format control. For detailed formatting information, please refer to our project's open-source repository: \href{\github}{\github}.
\label{app:prompts}
\subsection{Sticker Process}
\begin{figure*}[htbp]
    \centering
    \includegraphics[width=\linewidth]{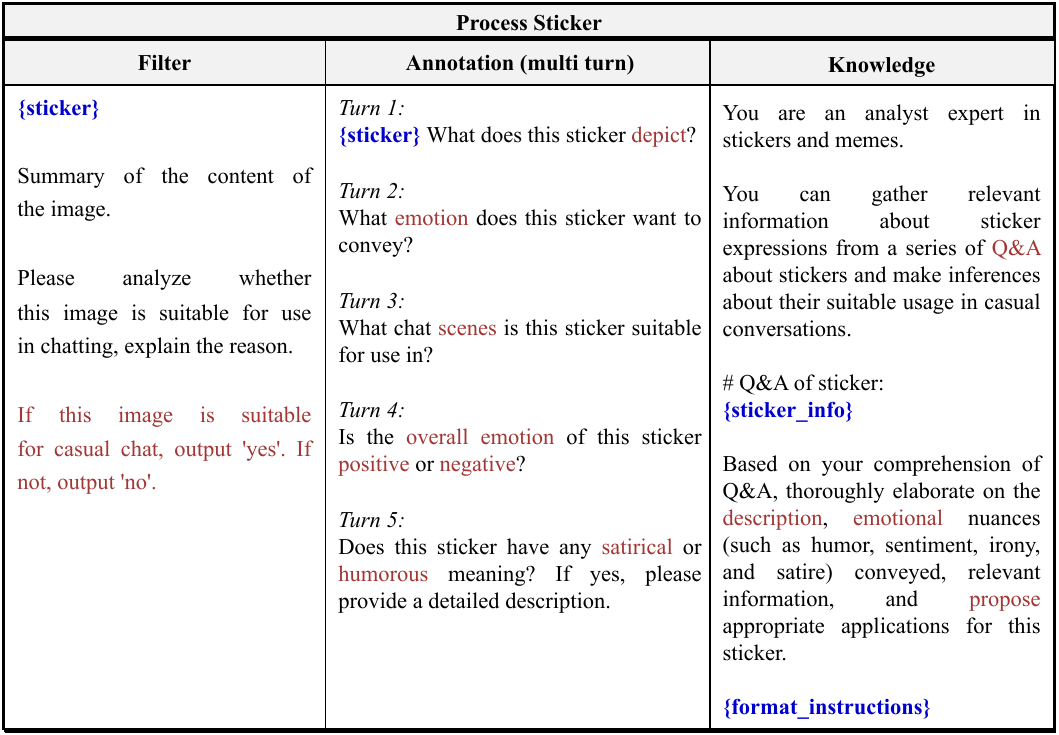}
    \caption{Prompt Templates for Process Sticker.}
    \label{prompt: llava}
\end{figure*}

Figure \ref{prompt: llava} shows the prompt of sticker process. For each sticker, we subject it to a process of filtering, annotation, and knowledge extraction, ultimately yielding three pieces of information: description, emotion and recommendation.

\subsection{Sticker Agent Chat}
\begin{figure*}[!htbp]
    \centering
    \includegraphics[width=0.75\linewidth]{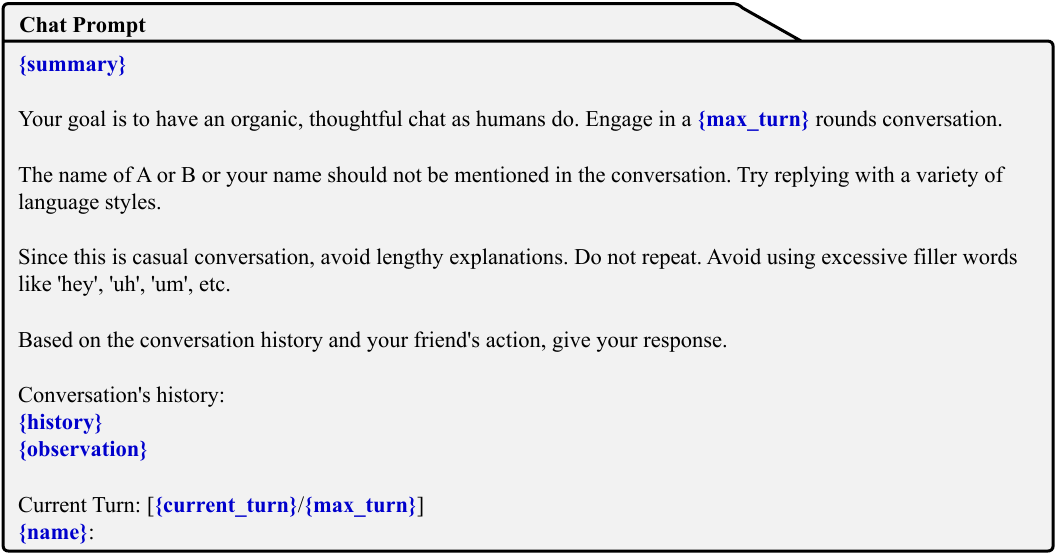}
    \caption{Prompt Template for Chat Prompt.}
    \label{chat_prompt}
\end{figure*}

\begin{figure*}[!htbp]
    \centering
    \includegraphics[width=0.75\linewidth]{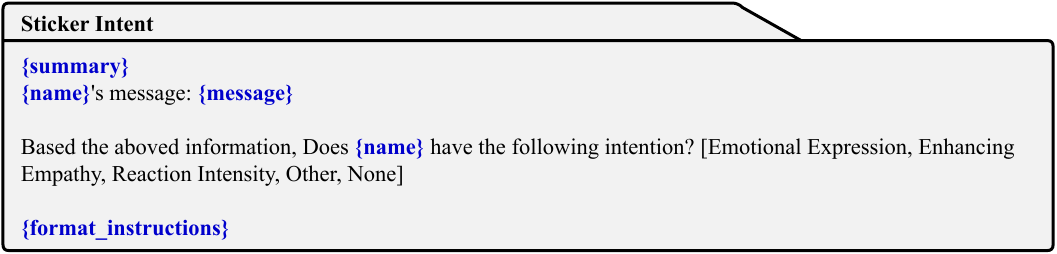}
    \caption{Prompt Template for Manager Agent.}
    \label{prompt:sticker_intent}
\end{figure*}

\begin{figure*}[!htbp]
    \centering
    \includegraphics[width=0.75\linewidth]{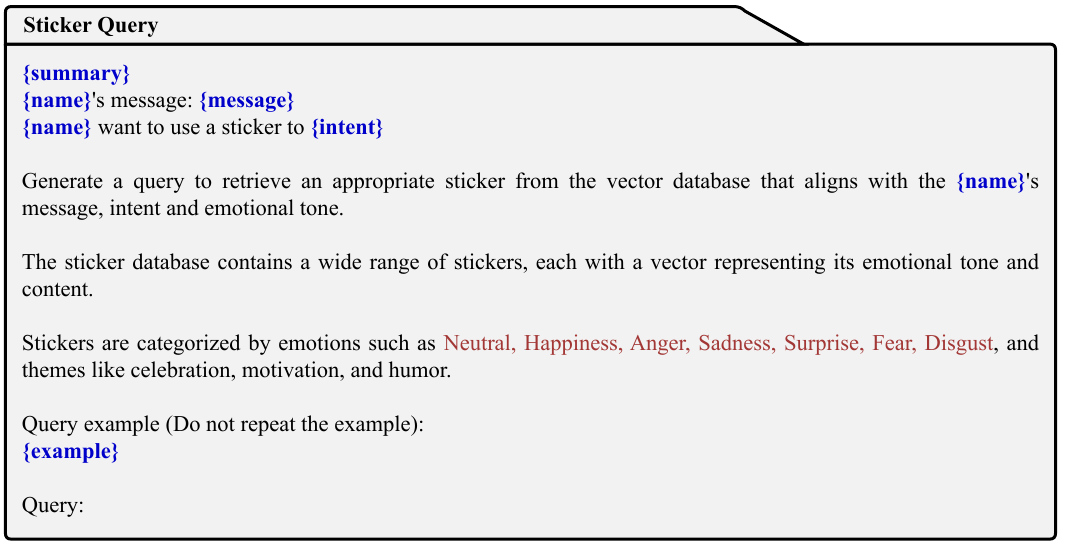}
    \caption{Prompt Template for Sticker Query.}
    \label{prompt:sticker_query}
\end{figure*}

\begin{figure*}[!htbp]
    \centering
    \includegraphics[width=0.75\linewidth]{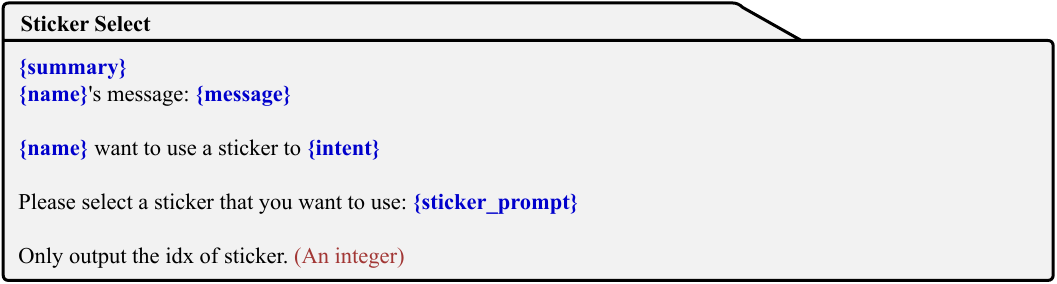}
    \caption{Prompt Template for Sticker Select.}
    \label{prompt:sticker_select}
\end{figure*}

Figure \ref{chat_prompt} shows the chat prompt, the summary encompasses the persona of the user or system, the current status, and the core traits generated based on memory. Figure \ref{prompt:sticker_intent}, \ref{prompt:sticker_query}, \ref{prompt:sticker_select} show the process of used sticker.

\subsection{Manager Agent}
\begin{figure*}[!htbp]
    \centering
    \includegraphics[width=0.75\linewidth]{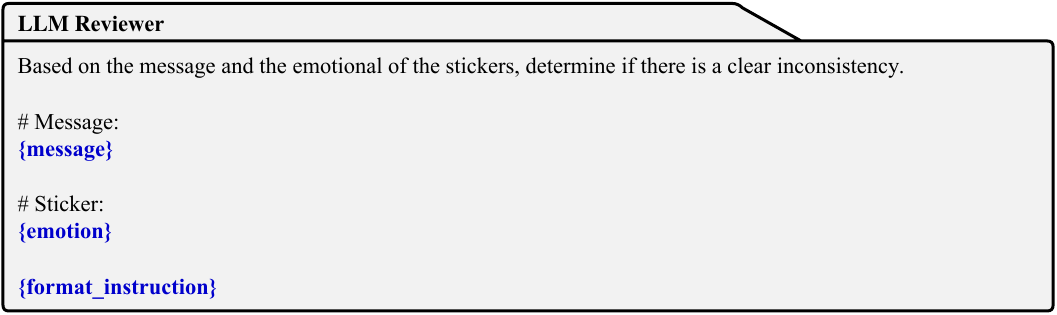}
    \caption{Prompt Template for Sticker Intent.}
    \label{prompt:llm_reviewer}
\end{figure*}

Figure \ref{prompt:llm_reviewer} shows the prompt of manager agent. By using various "\textbf{\textit{format instructions}}", this prompt can serve multiple purposes, including quality and consistency checks, among others.

\subsection{LLM-Based Scorer}
\begin{figure*}[ht]
    \centering
    \includegraphics[width=0.75\linewidth]{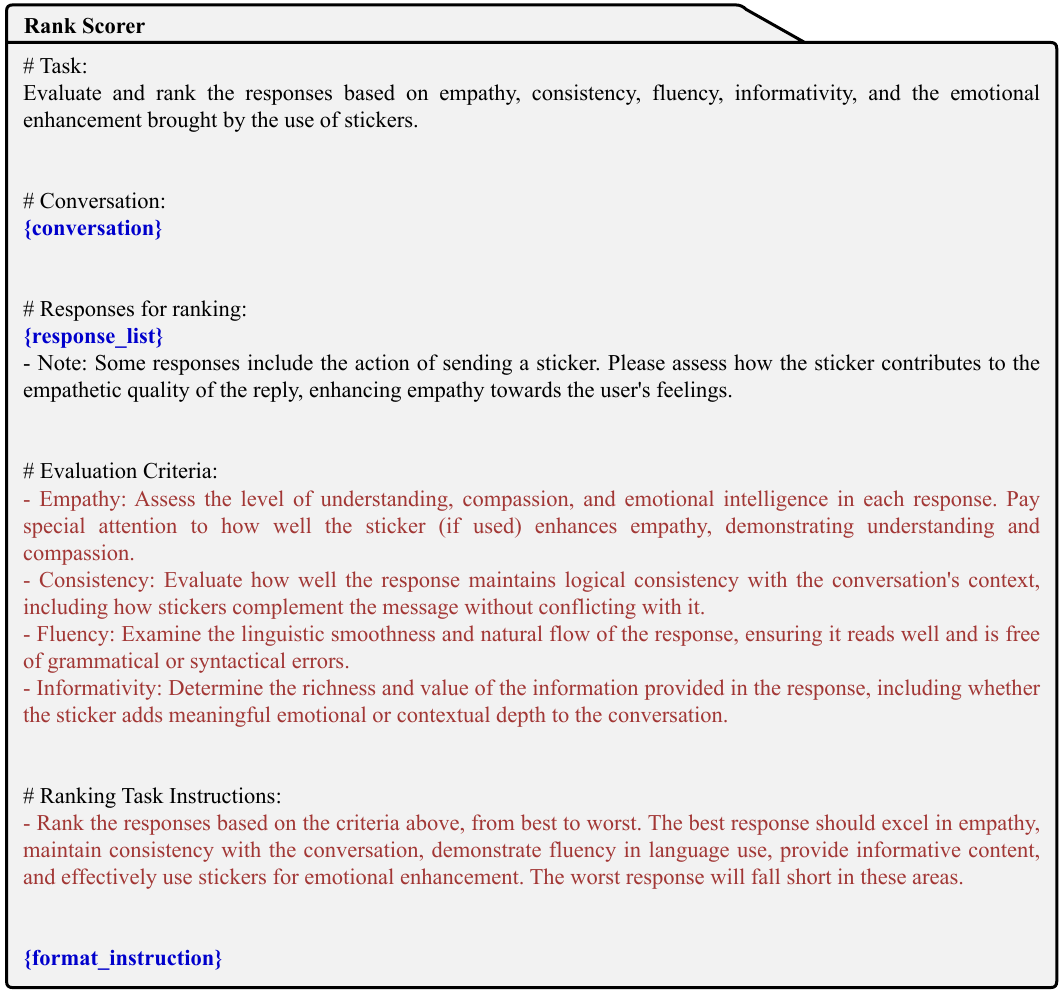}
    \caption{Prompt Template for Ranking score.}
    \label{ranking_scorer}
\end{figure*}

Figure \ref{ranking_scorer} illustrates the template for the preference rank scorer. This scorer integrates the outputs from all baseline models as well as \ourmodel, constituting the \textcolor[RGB]{0,0,204}{\rm{\{response\_list\}}}. It then proceeds to rank these responses and provide reasoning for the assigned rankings.

\begin{figure*}[htbp]
    \centering
    \includegraphics[width=0.75\linewidth]{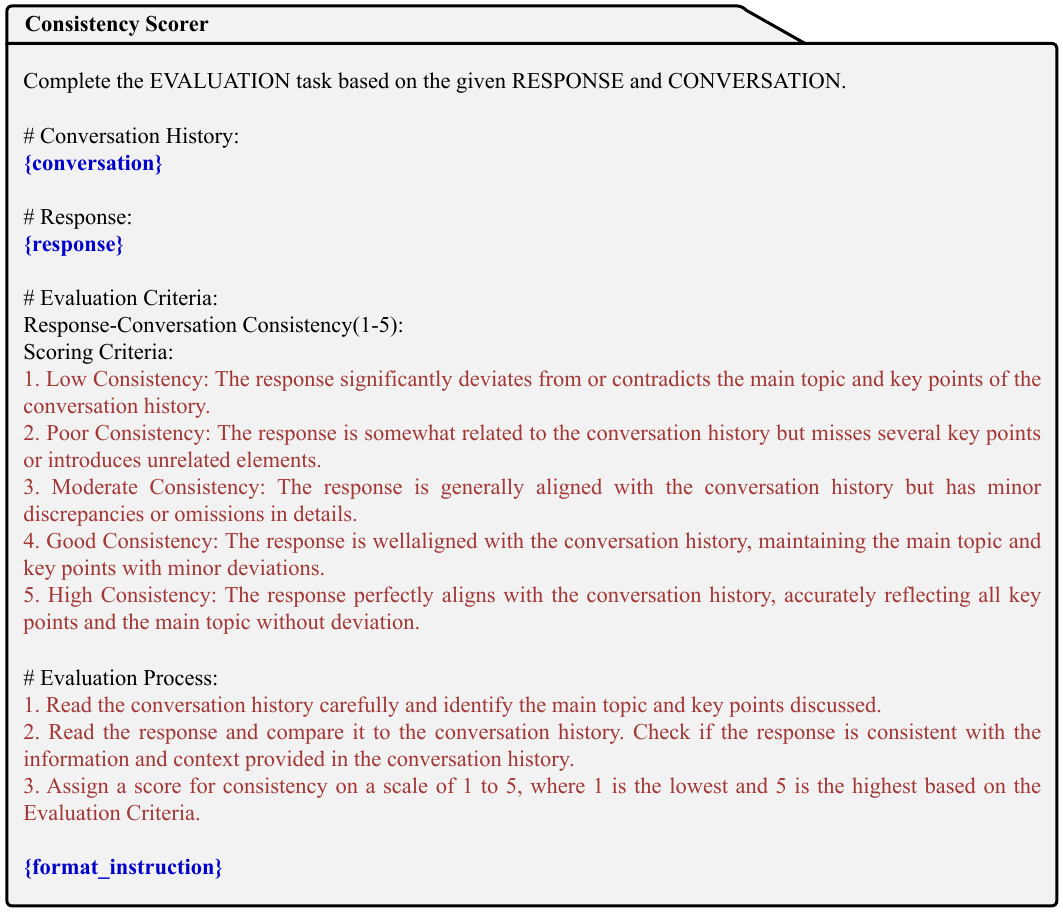}
    \caption{Prompt Template for Consistency Scorer.}
    \label{consistency_scorer}
\end{figure*}

Figures \ref{consistency_scorer} depict the templates for the consistency scorer. It is utilized for scoring purely textual responses.

\begin{figure*}[htbp]
    \centering
    \includegraphics[width=0.75\linewidth]{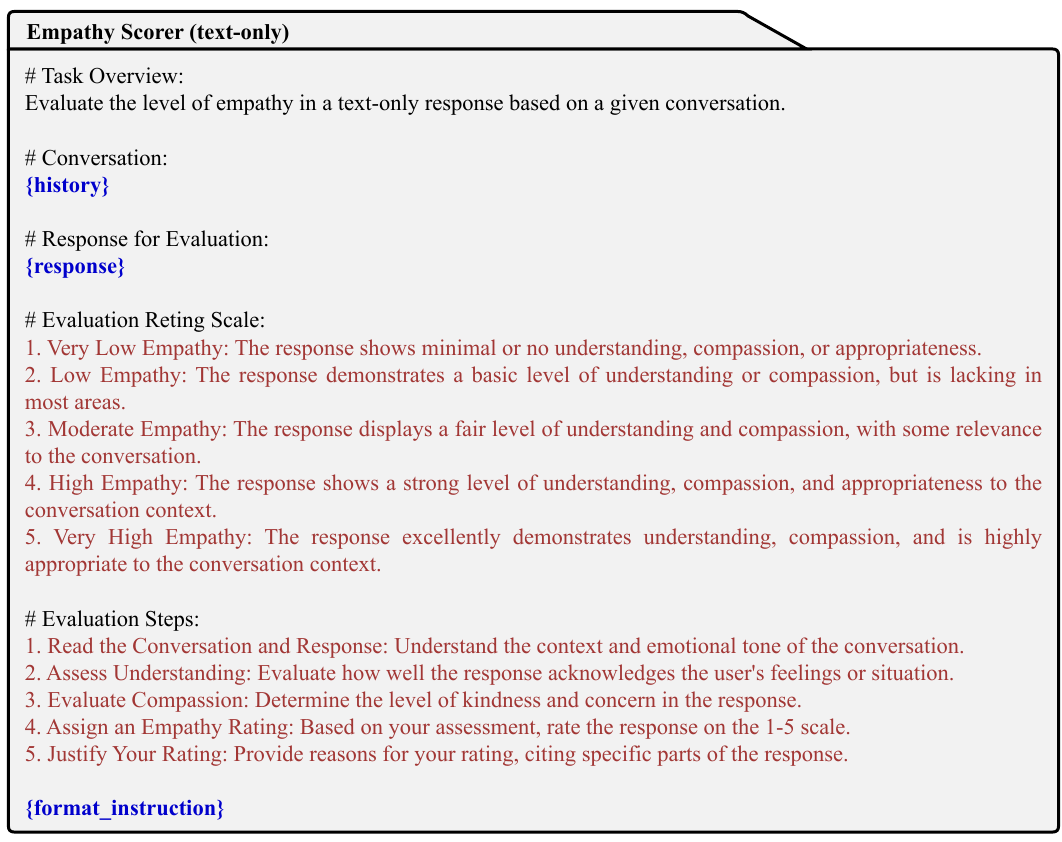}
    \caption{Prompt Template for Empathy Scorer (text only).}
    \label{empathy_scorer}
\end{figure*}

\begin{figure*}[htbp]
    \centering
    \includegraphics[width=0.75\linewidth]{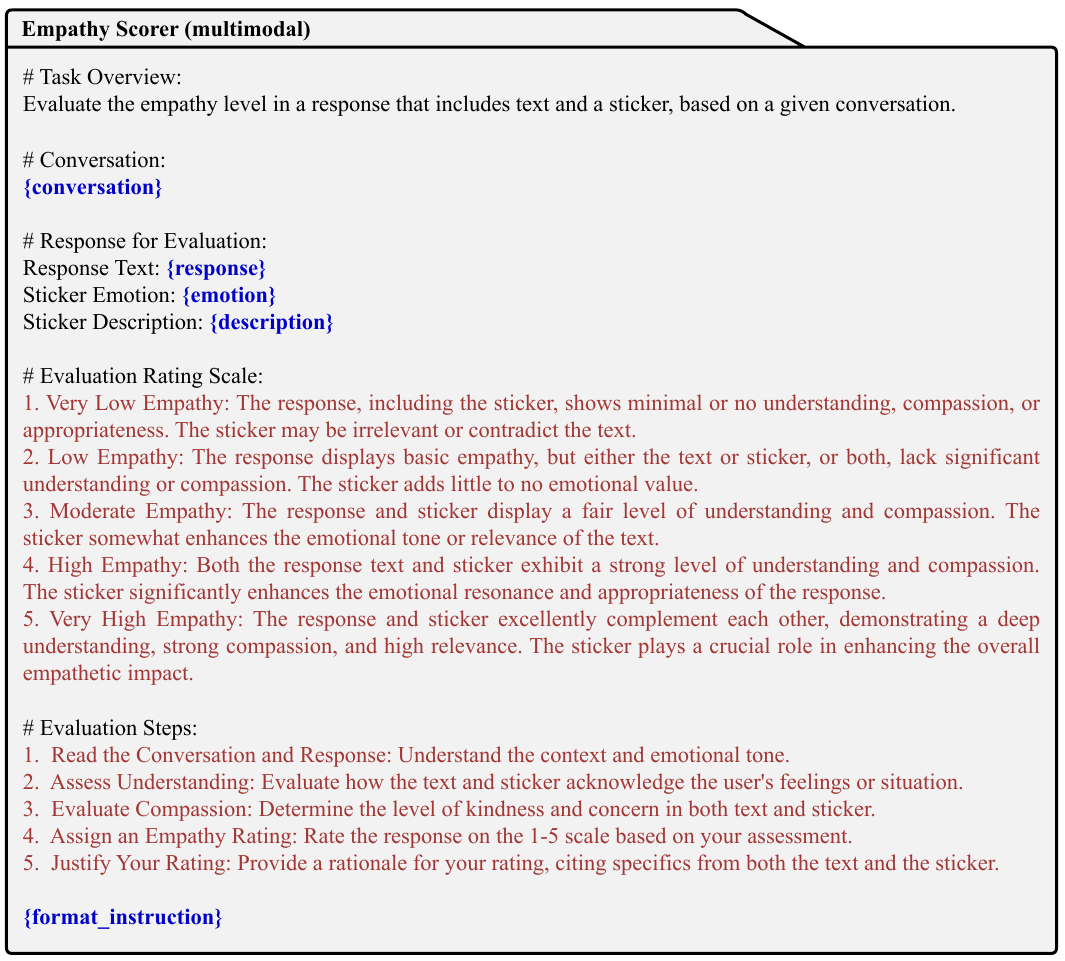}
    \caption{Prompt Template for Empathy Scorer (multimodal).}
    \label{empathy_scorer_sticker}
\end{figure*}

Figures \ref{empathy_scorer} and \ref{empathy_scorer_sticker} present the templates for the empathy scorer. The methodology for processing responses of different modalities aligns with the approach described for the consistency scorer.

\subsection{Joint Learning}

\begin{figure*}[htbp]
    \centering
    \includegraphics[width=0.75\linewidth]{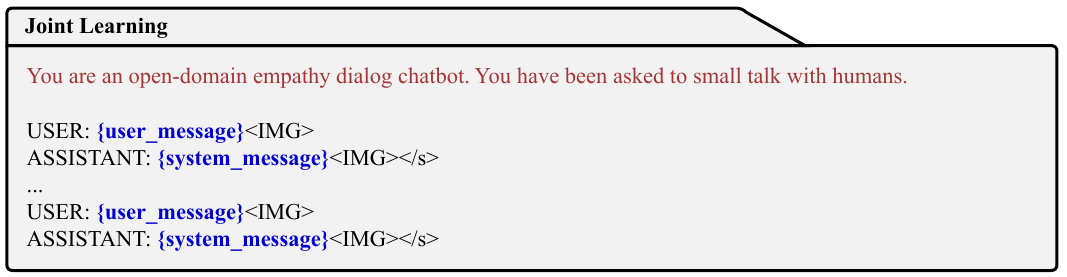}
    \caption{Prompt Template for Joint Learning.}
    \label{joint_learning}
\end{figure*}

Figure \ref{joint_learning} shows the prompt template for joint learning. We employed an instruction template similar to Vicuna v1.5. For every dialogue that uses a sticker, we append an <IMG> identifier.

\subsection{Tool Learning}
\begin{figure*}[htbp]
    \centering
    \includegraphics[width=0.75\linewidth]{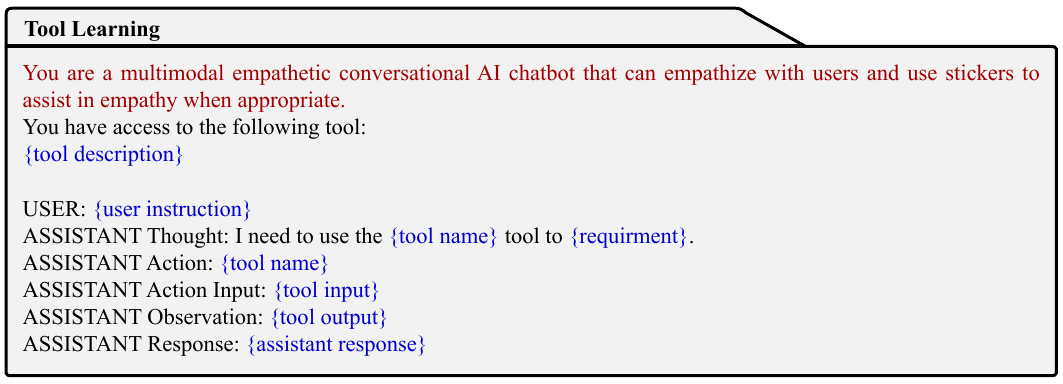}
    \caption{Prompt Template for Tool Learning.}
    \label{tool_learning}
\end{figure*}

Figure \ref{tool_learning} presents the instruction prompts for tool learning.
To instruct the LLM in utilizing the SD model for sticker generation, we transform the \dataset to tool format. When the model requires invoking the SD model, it outputs "\textbf{\textit{ASSISTENT Action}}" along with the corresponding prompt ("\textbf{\textit{ASSISTENT Action Input}}"). We then feed this prompt into the pre-configured SD model and relay the outcomes back to the LLM as "\textbf{\textit{ASSISTENT Observation}}".

\section{Costing}
\label{app: costing}
Extracting knowledge from labeled stickers in Llava using ChatGPT-3.5-turbo, which includes stickers in SER30K dataset, stickers generated by tool models and \ourmodel, costs approximately \$130.

The construction of 2000 user profiles for \agent with ChatGPT-3.5-turbo incurred a total cost of \$5.5.

Constructing the \dataset dataset with ChatGPT-3.5-turbo incurred a total cost of \$438.53, covering training, validation, and testing sets, at an average of \$0.03 per dialogue session.

According to our test set, evaluating metrics other than rank for each comparison model costs \$22. Combining all model responses with the evaluation Rank metric costs about \$60.

\label{sec:appendix}

\end{document}